\documentclass[10pt,twocolumn,letterpaper]{article}

\usepackage[pagenumbers]{cvpr} 

\usepackage[accsupp]{axessibility}  

\usepackage{multirow}

\usepackage{algpseudocode}
\usepackage{algorithm}
\usepackage{tabularx}
\usepackage{amsmath}
\usepackage{tabularray}
\makeatletter
\newcommand{\multiline}[1]{%
  \begin{tabularx}{\dimexpr\linewidth-\ALG@thistlm}[t]{@{}X@{}}
    #1
  \end{tabularx}
}
\makeatother

\DeclareMathOperator*{\argmin}{arg\,min}

\newcommand{\topk}[1]{\mathop{\mathrm{top}^{#1}}}

\def\tablescaler{0.85}

\definecolor{cvprblue}{rgb}{0.21,0.49,0.74}
\usepackage[pagebackref,breaklinks,colorlinks,allcolors=cvprblue]{hyperref}

\def\paperID{13853} 
\def\confName{CVPR}
\def\confYear{2025}

\title{Closed-Loop Supervised Fine-Tuning of Tokenized Traffic Models}

\author{Zhejun Zhang$^{1}$ \quad Peter Karkus$^{1}$ \quad Maximilian Igl$^{1}$ \\ Wenhao Ding$^{1}$ \quad Yuxiao Chen$^{1}$ \quad Boris Ivanovic$^{1}$ \quad Marco Pavone$^{1,2}$ \\
$^{1}$NVIDIA Research \quad $^{2}$Stanford University\\
{\tt\small \{zhejunz, pkarkus, migl, wenhaod, yuxiaoc, bivanovic, mpavone\}@nvidia.com} \\
{\tt\small pavone@stanford.edu}
}

\begin{document}
\maketitle

\begin{abstract}
Traffic simulation aims to learn a policy for traffic agents that, when unrolled in closed-loop, faithfully recovers the joint distribution of trajectories observed in the real world.
Inspired by large language models, tokenized multi-agent policies have recently become the state-of-the-art in traffic simulation.
However, they are typically trained through open-loop behavior cloning, and thus suffer from covariate shift when executed in closed-loop during simulation. 
In this work, we present Closest Among Top-K (CAT-K) rollouts, a simple yet effective closed-loop fine-tuning strategy to mitigate covariate shift. 
CAT-K fine-tuning only requires existing trajectory data, without reinforcement learning or generative adversarial imitation.
Concretely, CAT-K fine-tuning enables a small 7M-parameter tokenized traffic simulation policy to outperform a 102M-parameter model from the same model family, achieving the top spot on the Waymo Sim Agent Challenge leaderboard at the time of submission.
The code is available at \url{https://github.com/NVlabs/catk}.
\end{abstract}

\section{Introduction}
\label{sec:intro}

\begin{figure}[t]
  \centering
   \includegraphics[width=0.92\linewidth]{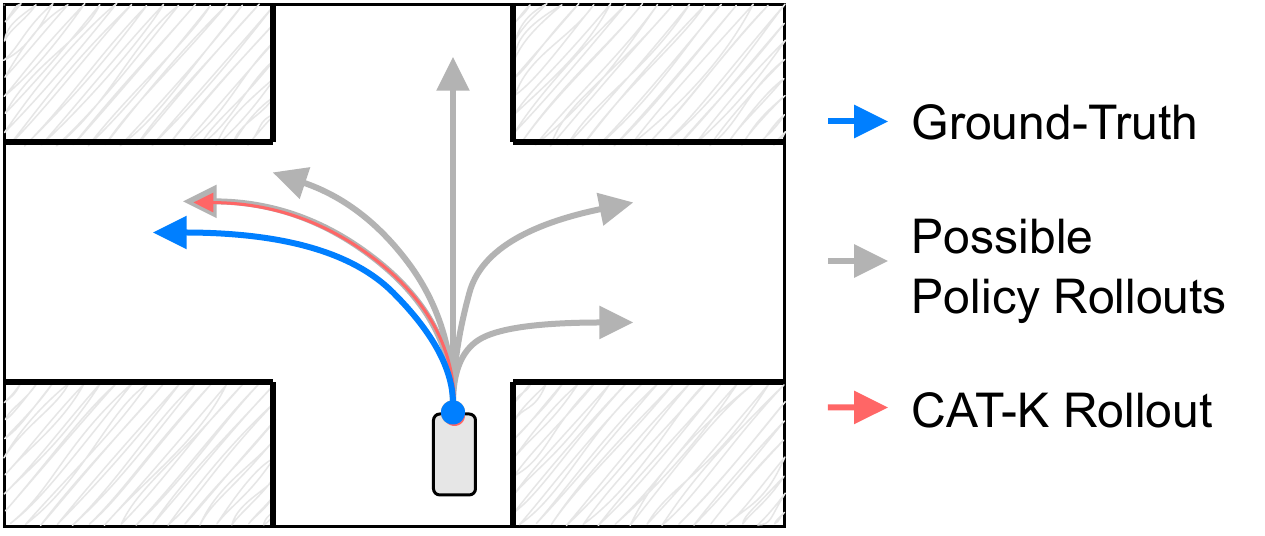}
   \vspace{-1.5ex}
   \caption{
   \textbf{Closest Among Top-K (CAT-K) rollouts.} 
   The key idea of our approach is to unroll the policy during fine-tuning in a way that visited states remain close to the GT. 
   At each time step, CAT-K first takes the top-K most likely action tokens according to the policy, then chooses the one leading to the state closest to the GT. 
   As a result, CAT-K rollouts follow the mode of the GT (e.g., turning left), while random or top-K rollouts can lead to large deviations (e.g., going straight or right). Since the policy is essentially trained to minimize the distance between the rollout states and the GT states, the GT-based supervision remains effective for CAT-K rollouts, but not for random or top-K rollouts.
   }
   \label{fig:teaser}
   \vspace{-1ex}
\end{figure}

Traffic modeling is a cornerstone of autonomous driving simulation and evaluation, typically formulated as learning a multi-agent policy that imitates the behavior of traffic participants in the real world. Given a set of historical agent trajectories and scene context (map, traffic light states, etc.), the policy generates actions for all simulated agents. The task gives rise to an imitation learning (IL) problem, with two key challenges: \emph{multimodality} and \emph{covariate shift}. 

Traffic agent behavior is highly multimodal, and faithfully recovering accurate behavior distributions is a key challenge in the field. 
Inspired by large language models~\cite{brown2020language,touvron2023llama}, recent works introduce next-token-prediction (NTP) models where the policy reduces to a classifier over a discrete set of trajectory tokens, which makes it easier to represent highly-multimodal distributions. 
Accordingly, the Waymo Open Sim Agent Challenge (WOSAC) leaderboard~\cite{montali2024waymo} is heavily populated by tokenized traffic models~\cite{wu2025smart,zhao2024kigras,zhou2024behaviorgpt,hu2025solving}. 

Covariate shift is a well-known challenge of IL arising from the gap between open-loop training and closed-loop deployment. When a model is trained on a fixed dataset of expert demonstrations, it can face a distribution mismatch between the states seen during training and those encountered during deployment, as small errors compound and lead to unseen states where the policy performs poorly.
A classic approach to tackle this problem is Dataset Aggregation (DAgger)~\cite{ross2011reduction}, which unrolls the policy and queries an expert to generate new demonstrations, but querying experts is not readily available for traffic simulation. 
Prior work has proposed closed-loop training using hand-crafted recovery controllers \cite{bansal2018chauffeurnet} or reinforcement learning (RL)~\cite{lu2023imitation,peng2025improving,zhang2023learning}. However, it is inherently difficult to design rewards with high behavioral realism or recovery controllers robust to divergent modes. Consequently, such approaches are not currently competitive on WOSAC realism metrics. 

\textbf{Contributions:} We introduce Closest Among Top-K (CAT-K) rollouts, a simple yet highly efficient fine-tuning strategy to address the open-loop to closed-loop gap. The key idea, illustrated in \cref{fig:teaser}, is to unroll the multimodal policy during training in a way that the policy-visited states remain close to the ground-truth (GT) demonstration. CAT-K achieves this by first finding the K most likely modes of the policy and then choosing the mode closest to the GT. At inference time, actions are sampled from the policy according to the predicted likelihoods. During training, however, random sampling would lead to large deviations from the demonstrations, making them invalid and degrading the final policy performance. Our CAT-K rollout strategy balances being on-policy and staying close to GT demonstrations, such that they remain a valid supervision signal.


Experiments on the Waymo Open Motion Dataset (WOMD) demonstrate the efficacy of CAT-K. Notably, fine-tuning the SMART-7M~\cite{wu2025smart} next-token-prediction traffic model enables it to outperform the 14x larger State-of-The-Art (SoTA) SMART-102M from the same model family, achieving the \#1 spot on the public WOSAC leaderboard at the time of submission.
To further demonstrate the potential of employing CAT-K fine-tuning for different tasks and policy representations, we apply it to an ego-motion planning task using a Gaussian Mixture Model (GMM) policy, yielding significant gains in closed-loop behavior, reducing collisions by 25.7\% and off-road driving by 33.9\%.



\section{Related work}
\label{sec:related_work}

\subsection{Traffic simulation}
Prior work explored various architectures for traffic models, including conditional variational autoencoders~\cite{suo2021trafficsim,xu2023bits,igl2023hierarchical}, motion forecasting Transformers~\cite{zhou2023query,girgis2021latent,ngiam2021scene,shi2024mtr++}, and diffusion models \cite{pronovost2023scenario,zhong2023guided,zhong2023language,jiang2023motiondiffuser,huang2024versatile,lu2024scenecontrol}.
However, long-term stability is an open challenge for these models due to covariate shift when transitioning from open-loop training to closed-loop deployment.
Various ways have been proposed to mitigate covariate shift.
BITS~\cite{xu2023bits} and Symphony~\cite{igl2022symphony} introduce hierarchy, with high-level intent and low-level behavior prediction.
TrafficBots~\cite{zhang2023trafficbots, zhang2024trafficbots} incorporates configurable behaviors through destination and personality.
Most diffusion models use guiding to generate rule-abiding behavior; however, they often struggle with computational efficiency and long horizons. 
The latest advances come from NTP models, which predict the next action as a token, e.g., Trajeglish~\cite{philion2024trajeglish}, GUMP~\cite{hu2025solving}, KiGRAS~\cite{zhao2024kigras}, MotionLM~\cite{seff2023motionlm}, and SMART~\cite{wu2025smart}.
Notably, SMART is the current SoTA on the WOSAC leaderboard. 
In addition to the strong scalability and flexibility, NTP models also show better closed-loop stability than regression-based models, thanks to their discrete action space.
However, achieving generalization and reducing compounding errors continue to be challenges.

\subsection{Data augmentation for behavior cloning}

Data augmentation is a simple yet effective way to improve the generalizability of traffic simulation models.
ChauffeurNet~\cite{bansal2018chauffeurnet} showed that carefully perturbing the vehicle trajectory and designing a recovery trajectory could alleviate the covariate shift suffered by behavior cloning (BC).
However, this technique is difficult to apply to traffic simulation with complicated scenarios, including pedestrians and cyclists; and adding handcrafted recovery trajectories may negatively impact behavioral realism. 
Recently, NTP works such as Trajeglish~\cite{philion2024trajeglish} and SMART~\cite{wu2025smart} have explored similar ideas by using noisy tokenization to perturb trajectories during training, but their data augmentation did not lead to significant improvements in performance.
Our method is related to Trajeglish's noisy tokenization, but importantly, instead of blindly sampling tokens close to the GT without considering the policy, our CAT-K rollout selects the token from the most likely K tokens predicted by the policy that is closest to the GT.
It can be considered as a closed-loop variants of the ``winner-takes-all'' training strategy~\cite{makansi2019overcoming} widely applied to open-loop motion prediction models~\cite{zhang2024real,ngiam2021scene,shi2022motion}.


\subsection{Closed-loop fine-tuning}

As mentioned above, covariate shift is a major challenge faced by traffic simulation models, since most of them are trained in open-loop and evaluated in closed-loop. Even with data augmentation, the issue is not fully resolved as the augmented noisy data does not reflect the compounded error during closed-loop rollout.
Therefore, some existing works explored the use of closed-loop fine-tuning. 

As a classic remedy for the covariate shift, DAgger has been applied to end-to-end driving~\cite{zhang2016query,prakash2020exploring,zhang2021roach}, yet its application in traffic simulation is limited, as it requires interactive demonstrations from human drivers or an expert policy.
As an expert-free variant of DAgger, Data As Demonstrator (DAD)~\cite{venkatraman2015improving} obtains recovery actions using a pair of trajectories: a forward simulation (i.e., rollout) trajectory and a GT trajectory.
At each time step, the policy is trained to return to the GT next state from the simulated current state.
However, DAD fails when the rollout deviates from the GT trajectory.
Our CAT-K rollout directly addresses this problem, enabling the use of DAD for closed-loop fine-tuning of traffic simulation and ego-vehicle policies.

Another popular approach is to use RL. BC is combined with RL to improve the robustness of a policy in~\cite{lu2023imitation}. Yet, it also exposes RL's weakness in improving realism as it is difficult to handcraft a reward that promotes realism. Its follow-up work~\cite{peng2025improving} learns a joint traffic model capable of rolling out the entire scenario by itself. The reward/loss are typically handcrafted, and can be distilled from explicit traffic rules~\cite{zhang2023learning}. With human preference data, RL from human feedback (RLHF) has also been applied to traffic model training for better user alignment~\cite{huang2024gen,cao2024reinforcement}. 
Finally, Symphony \cite{igl2022symphony} adds a Generative Adversarial Imitation Learning (GAIL) loss to encourage the rollout states to stay in distribution; however, the well-known issue of training stability and discriminator overfitting remains a challenge.

\section{Background}

\subsection{Problem formulation}

A multi-agent traffic simulation policy can be typically formulated as $\pi_\theta(\mathbf{a}_t|\mathbf{h}_t, \mathcal{M})$, where $\theta$ denotes the trainable model parameters, $\mathbf{h}_t=\mathbf{s}_{t-H:t}$ is the state history of length $H$, $\mathcal{M}$ is the context, including for example high-definition (HD) maps and traffic light states, $t$ is the current time step, $\mathbf{a}_t=[a_t^1, ..., a_t^N]$ and $\mathbf{s}_t=[s_t^1, ..., s_t^N]$ are respectively the actions and states of $N$ agents at the current time step.
The dimensions of actions $a$ and states $s$ are respectively denoted as $D_a$ and $D_s$, i.e., $a\in\mathbb{R}^{D_a}$ and $s\in\mathbb{R}^{D_s}$.
From the current states and actions at step $t$, the next states are computed using the per-agent forward dynamics $\mathbf{s}_{t+1}=f(\mathbf{s}_{t}, \mathbf{a}_{t})=\left[f({s}_{t}^i, {a}_{t}^i)\right]_{i=1}^N$. We assume that $f({s}_{t}^i, {a}_{t}^i)$ is deterministic, and can be queried during training, which is the case for traffic simulation. Extensions to stochastic dynamics would be possible in future work. 
We define a rollout of $T$ steps starting at $t=0$ as a sequence of states $\mathbf{s}_{0:T}=[\mathbf{s}_{0}, \dots, \mathbf{s}_{T}]$, while the GT trajectories of all agents are denoted as $\hat{\mathbf{s}}_{0:T}$.
For the training, we are given a dataset $\mathcal{D}=\{\mathbf{\hat{s}}_{0:T}^{j},\mathcal{M}^j\}_{j=1}^{|\mathcal{D}|}$ of such real-world trajectories that we want to emulate with their corresponding contexts. 

\subsection{Next Token Prediction (NTP) policies}

NTP policies, such as SMART~\cite{wu2025smart} and Trajeglish~\cite{philion2024trajeglish}, are parameterized as a probability distribution over a vocabulary of action tokens denoted as $V=\{x_c \mid c = 1, 2, \dots, |V| \}$, where $|V|$ is the size of the vocabulary, $x_c \in \mathbb{R}^{D_a}$ are template actions and $c \in \mathbb{N}$ is the token index.
Hence, an autoregressive NTP policy for traffic simulation can equivalently be written as an agent-factorized categorical distribution at each timestep $t$, i.e.,
\begin{equation*}
    \pi_\theta(\mathbf{c}_t \mid \mathbf{h}_{t}, \mathcal{M}) = \prod\limits_{i=1}^N \pi_\theta(c_t^i \mid \mathbf{h}_{t}, \mathcal{M}) = \prod\limits_{i=1}^N\operatorname{Cat}(c_t^i),
\end{equation*}
where $\operatorname{Cat}(c^i_t)$ is the categorical distribution over the action token index for agent $i$ (and not to be confused with our method CAT-K).
Given the sampled output $\mathbf{c}_t=[c_t^1, ..., c_t^N]$, the actions $\mathbf{a}_t=[ x_{c_t^1}, ..., x_{c_t^N} ]$ are obtained using the token vocabulary $V$.

\section{Method}

Two key challenges in learning a policy from real-world trajectories are the \emph{multimodal} nature of the trajectory distribution and the problem of \emph{covariate shift} when policies are trained open-loop, resulting in a distribution mismatch between expert states seen during training and states visited during policy deployment.
Covariate shift can be overcome by closed-loop training, i.e., by training on trajectories sampled from the learned policy.
However, this requires the generation of expert actions (or other notions of optimality) to be used as training targets along those trajectories~\cite{ross2011reduction}. 
Querying a human expert is infeasible at scale, 
RL-based methods require hard-to-define rewards, and methods such as GAIL~\cite{ho2016generative} are prone to mode collapse.
Consequently, they fail to achieve good performance in the WOSAC challenge.

An alternative strategy for generating ``expert" actions is to construct recovery actions that bring the agent back to the available GT trajectory.
However, this is complicated by the multimodal nature of the data, as the available GT trajectory might not be a valid recovery target for the generated trajectory. For example, as shown in \cref{fig:teaser}, the GT trajectory $\hat{s}_{0:T}$ turns left at the intersection, while the sampled trajectory $s_{0:T}\sim\pi_\theta$ might go straight or turn right. 
As a result, while some SoTA traffic models~\cite{philion2024trajeglish,wu2025smart} augment the training data with recovery actions to reduce the covariate shift, they do so only from states that were reached by injecting small amounts of noise into the GT trajectory. 
This does guarantee that the GT trajectory remains a valid recovery target, but it \emph{completely ignores the learned policy} and the state distribution induced by it.
Instead, our method, \emph{Closest Among Top-K} (CAT-K) rollout, informs the sampling process by the learned policy, but biases it towards the GT trajectory to guarantee the validity of the recovery actions. 
While simple to implement, it yields significant performance improvements in our experiments. 

\begin{algorithm}[t]
    \caption{CAT-K fine-tuning}
    \label{alg:catk}
\begin{algorithmic}[1]
    \State \textbf{Input}: Policy $\pi_\theta$, action token vocabulary $V$, dataset $\mathcal{D}$
    \State \multiline{Pre-train $\pi_\theta(\mathbf{c}_t\mid \hat{\mathbf{h}}_t, \mathcal{M})$ with BC until convergence}
    \Repeat \Comment{Closed-loop supervised fine-tuning}
        \State \multiline{Sample a traffic scenario $\{\hat{\mathbf{s}}_{0:T},\mathcal{M}\}$}
        \State \multiline{Init rollout state $\mathbf{s}_0=\hat{\mathbf{s}}_0$ \Comment{CAT-K Rollout}}
        \For{$t$ in $[0, \dots, T-1]$} \Comment{$T$ steps}
            \For{$i$ in $[1, \dots, N]$} \Comment{$N$ agents}
                \State \multiline{Get action index for rollout $c^i_t$. (Eq.~\ref{eq:catk})}
                \State \multiline{Get next rollout state $s^i_{t+1}$. (Eq.~\ref{eq:catk_update_state})}
                \State \multiline{Compute target $\hat{c}^i_{t}$. (Eq.~\ref{eq:tokenize}) } 
            \EndFor
        \EndFor
        \State \multiline{Update $\theta$ by minimizing $\mathcal{L}_\theta(\mathbf{s}_{0:T}, \hat{\mathbf{c}}_{1:T}, \mathcal{M})$. (Eq.~\ref{eq:cross_entropy}) }
    \Until{convergence}
\end{algorithmic}
\end{algorithm}

\begin{figure*}[t]
    \centering
    \begin{subfigure}[t]{0.265\textwidth}
        \centering
        \includegraphics[width=\textwidth]{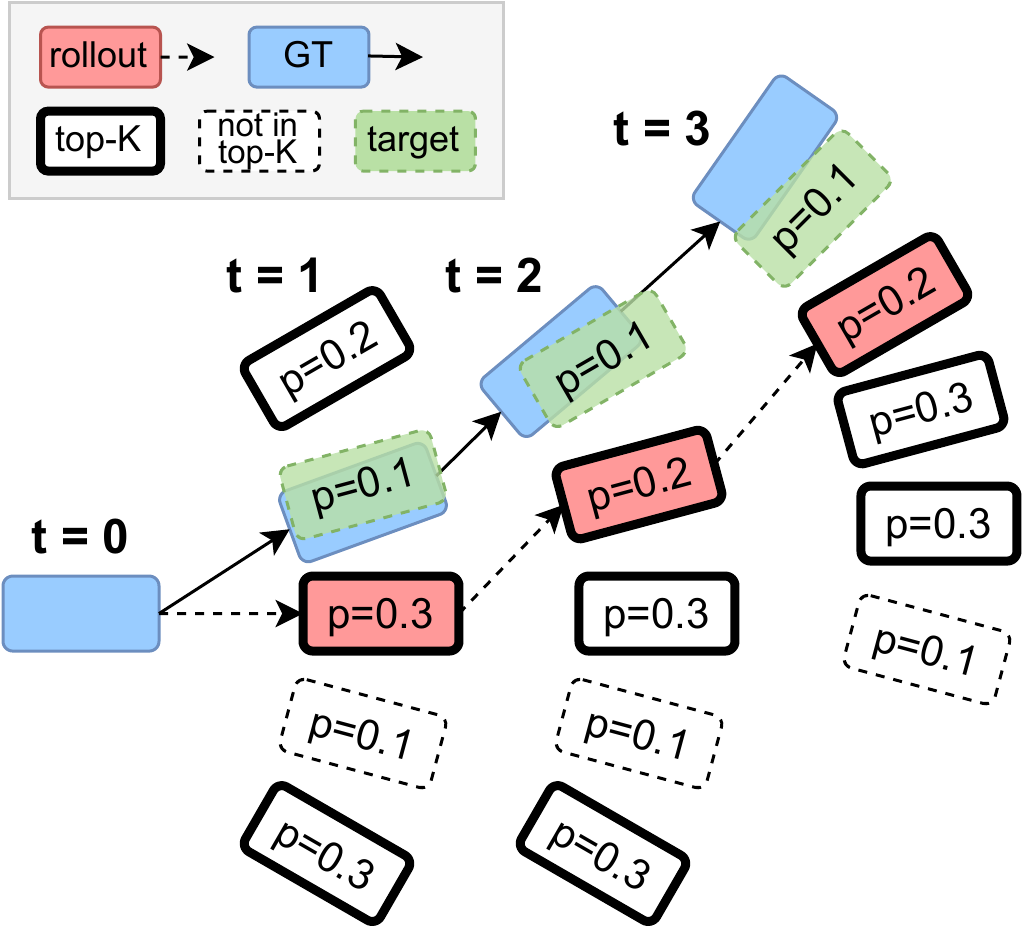}
        \caption{CAT-K rollout}
        \label{fig:catk_rollout}
    \end{subfigure}
    \hfill
    \begin{subfigure}[t]{0.235\textwidth}
        \centering
        \includegraphics[width=\textwidth]{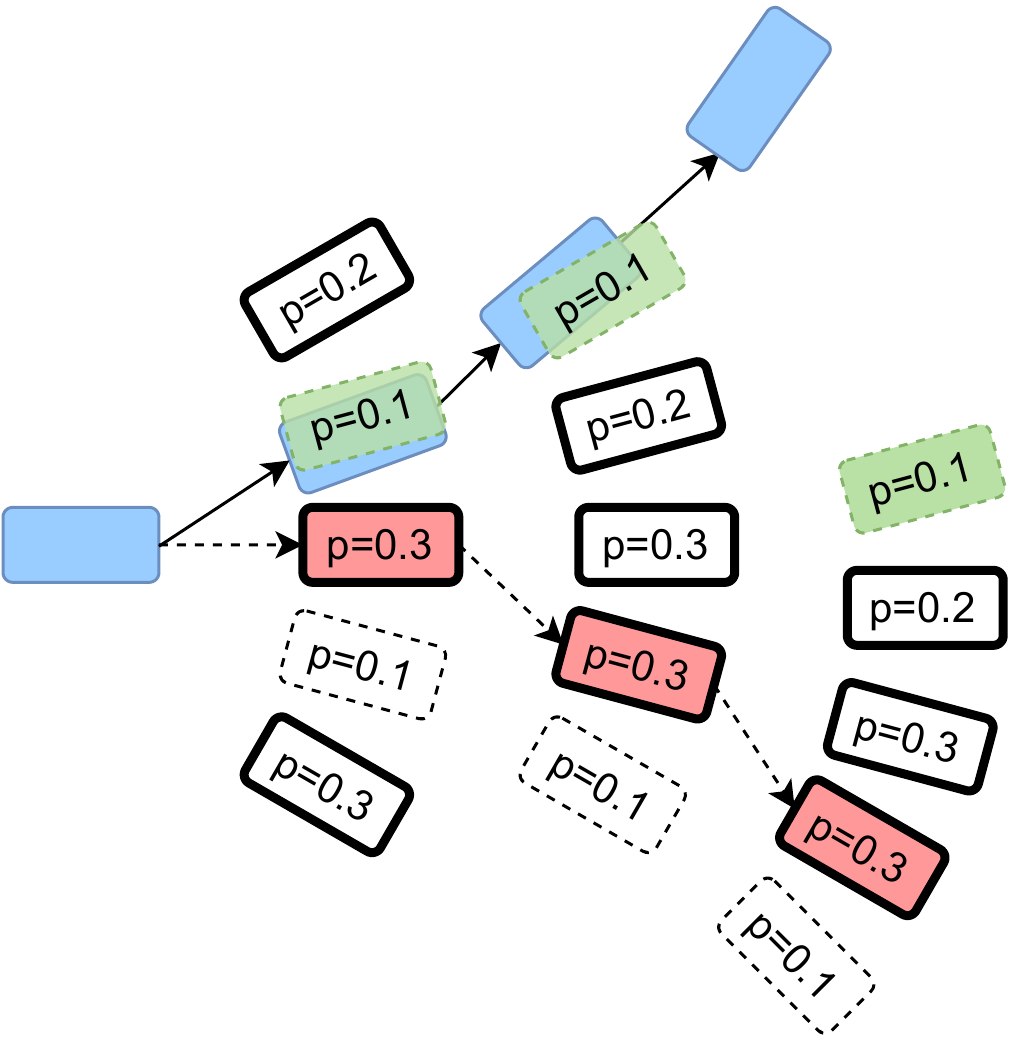}
        \caption{Top-K sampling}
        \label{fig:topk_rollout}
    \end{subfigure}
    \hfill
    \begin{subfigure}[t]{0.23\textwidth}
        \centering
        \includegraphics[width=\textwidth]{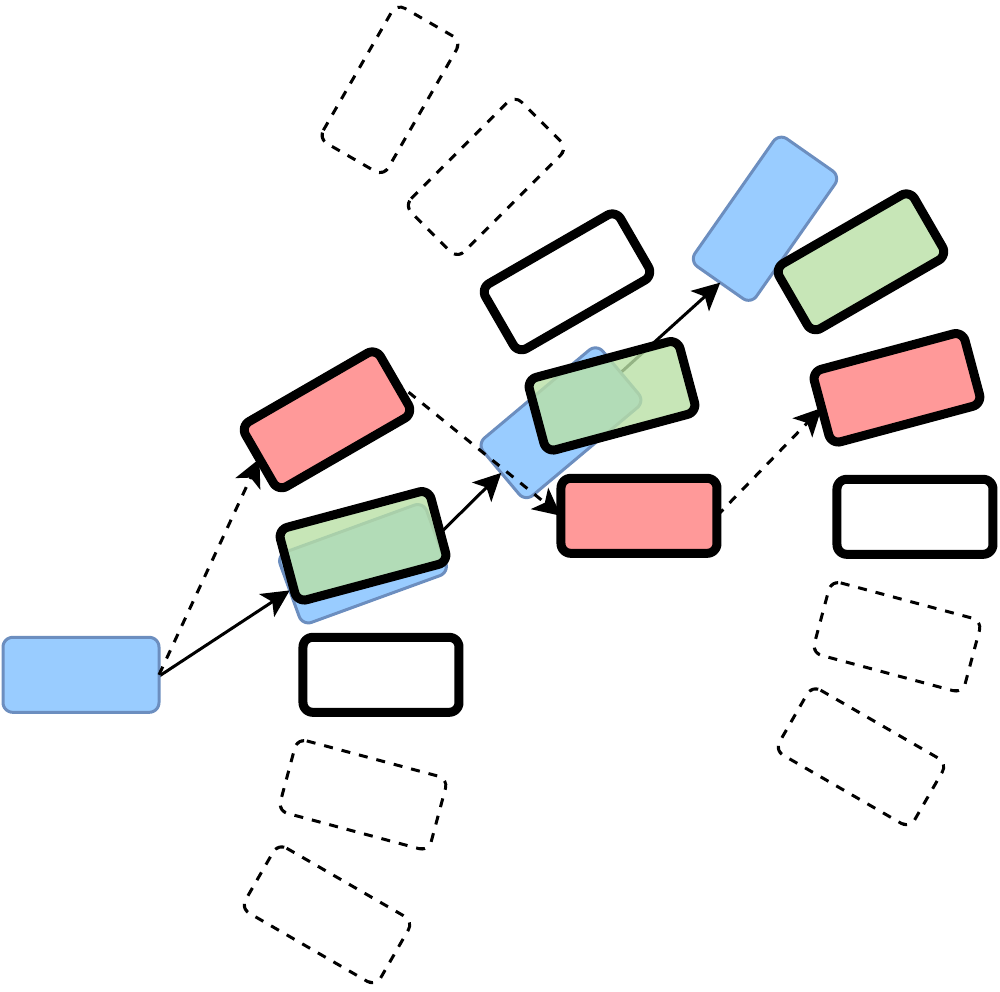}
        \caption{Trajeglish noisy tokenization}
        \label{fig:trajeglish_rollout}
    \end{subfigure}
    \hfill
    \begin{subfigure}[t]{0.23\textwidth}
        \centering
        \includegraphics[width=\textwidth]{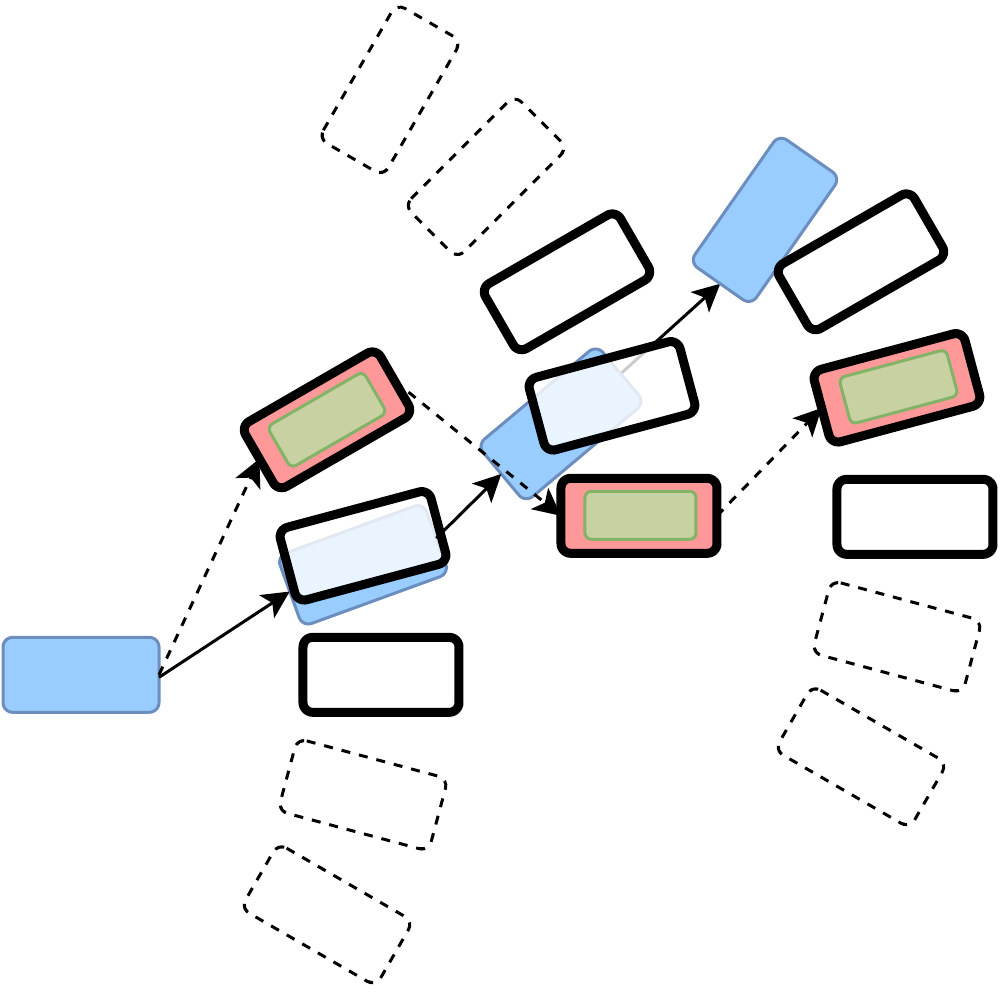}
        \caption{SMART trajectory perturbation}
        \label{fig:smart_rollout}
    \end{subfigure}
    \vspace{-1.5ex}
    \caption{\textbf{Schematic comparison of CAT-K rollout, top-K sampling, and data augmentation techniques of Trajeglish and SMART.}
    In this example, the token vocabulary has a size of 5.
    We rollout three steps from $t=0$ to $t=3$.
    For CAT-K rollout and top-K sampling, the top-K is w.r.t the probabilities $p$ of tokens predicted by the policy.
    For the data augmentations used by Trajeglish and SMART, the policy is unavailable, and the top-K selection is based on the negative distances between tokens and  GT.
    }
    \label{fig:catk_comparison}
    \vspace{-1.5ex}
\end{figure*}

\subsection{Closest Among Top-K (CAT-K) rollout}

To facilitate the formulation, we define a $\topk{K}$ operator:
\begin{equation*}
    \{\xi_1, \dots, \xi_K\} = \topk{K} \limits_{c \in \{1, \dots, |V|\}} (\operatorname{Cat}(c)),
\end{equation*}
where $\operatorname{Cat}: \mathbb{N} \to \mathbb{R}$ is the probability density of a categorical distribution on the vocabulary index, and $\{\xi_1, \dots, \xi_K\}$ are the $K$ most probable indices.
The $\topk{K}$ operator can be considered as a variation of the $\arg\max$ operator that returns multiple indices, with $\topk{1}$ equivalent to $\arg\max$.

At time step $t$, the policy $\pi_\theta(\mathbf{c}_t \mid \mathbf{h}_{t}, \mathcal{M})$ outputs independent categorical distributions over the token vocabulary for each agent.
Our method, CAT-K rollout, deterministically rolls out the policy by selecting, at each time step and for each agent, the one action among the top-K likeliest according to $\pi_\theta$ that brings the agent closest to the GT next state.
Using a distance metric $d(\cdot, \cdot)$ on the states, this is formally expressed as follows:
\begin{align}
    c^i_t &= \argmin\limits_{c\in \{ \xi_1,\dots,\xi_K \} } d\left( f(s^i_t, x_c), \hat{s}^i_{t+1} \right), \label{eq:catk} \\
     \{\xi_1, \dots, \xi_K\} &= \topk{K} \limits_{c^i_t \in \{1, \dots, |V|\}} \left[\pi(c_t^i|\mathbf{h}_t, \mathcal{M})\right], 
\end{align}
where $c^i_t$ is the action token indices of agent $i$ at step $t$ for the CAT-K rollout, and $\{\xi_1, \dots, \xi_K\}$ are the top-K likeliest token index according to the policy.
Given $c^i_t$, the next state is obtained using the vocabulary $V$ and the dynamics as:
\begin{equation}
    s^i_{t+1} = f(s^i_{t}, x_{c^i_t}) \label{eq:catk_update_state}.
\end{equation}
These rollout states will be used as the input $\mathbf{h}$ to the policy at the next time step.
By doing this sequentially from $t=0$ to $t=T-1$ and repeating for all $N$ agents, we obtain the CAT-K rollout trajectories $\mathbf{s}_{0:T}$.


\subsection{Closed-loop supervised fine-tuning}

Given the CAT-K rollout trajectory $\mathbf{s}_{0:T}$, we can apply the idea of DAD~\cite{venkatraman2015improving} and construct the recovery action indices $\hat{\mathbf{c}}_t$ from the GT trajectories $\hat{\mathbf{s}}_{0:T}$ by finding the action token that brings each agent closest to its original trajectory:
\begin{equation}
    \hat{c}_t^i = \argmin\limits_{c \in \{ 1,\dots, |V| \}} 
    d \left( f(s^i_t,x_c), \hat s^i_{t+1} \right).
    \label{eq:tokenize}
\end{equation}
Given these indices $\hat{\mathbf{c}}_t$, the NTP policy is trained using the cross-entropy loss
\begin{equation}
    \mathcal{L}_\theta =-\frac{1}{NT}\sum_{i=1}^N \sum_{t=0}^{T-1} \log \pi_\theta \left(\hat{c}_t^i \mid \mathbf{h}_{t}, \mathcal{M} \right).
    \label{eq:cross_entropy}
\end{equation}
Since CAT-K rollout is effective only when the top-K rollouts of the policy cover the GT mode, we adopt a two-stage training procedure, summarized in \cref{alg:catk}. First, we obtain a reasonably well-trained policy through BC pre-training, then fine-tune it using CAT-K rollouts.

\subsection{Comparison to previous methods}
\label{sec:comparison_prev_methods}

In \cref{fig:catk_comparison}, we compare our method with previous data augmentation approaches that 
alleviate the covariate shift for NTP traffic simulation policies.

\textbf{Top-K sampling} is a common approach used by NTP models for generating sequences \emph{during inference} as it improves the sample quality.
However, it is unsuitable for generating trajectories during training as it does not consider the distance to the GT trajectory, and hence the validity of generated recovery actions (c.f. \cref{fig:topk_rollout}).
This can be partially addressed by post-hoc filtering of trajectories that deviated too far from the GT, but the resulting method is sample-inefficient, as most rollouts will be discarded, and the distance threshold hyperparameter is difficult to tune, since its optimal value varies across scenarios (e.g., high-speed vs. low-speed situations).
In contrast, CAT-K rollout (c.f. \cref{fig:catk_rollout}) is sample efficient by choosing the closest among top-K actions at each timestep and also removes the need to tune a distance-based hyperparameter. 
As shown in \cref{table:ablation}, our hyperparameter $K$ provides strong results across a large range of values and is hence much easier to tune.

Instead of sampling top-K trajectories from the policy, current SoTA methods such as Trajeglish \citep{philion2024trajeglish} and SMART \citep{wu2025smart} use forms of \textbf{trajectory noising} to address the issue of covariate shift.
All trajectory noising approaches rely on the injection of small perturbations into the tokenization of the GT trajectory, but implementations can vary from each other in details.
For example, in Trajeglish, the likelihood of each noised token is a function of the resulting distance to the ground-truth trajectory, i.e. $q_t^i(c) \sim \exp(-d(f(s^i_t, x_c), \hat{s}^i_{t+1})/\tau)$, while for SMART, tokens are sampled uniformly from those $K$ tokens with the highest likelihood $q_t^i(c)$. 
Additionally, Trajeglish, similar to our methods, uses the DAD~\cite{venkatraman2015improving} recovery target that finds the token that would bring the agent back to the GT trajectory (c.f. \cref{fig:trajeglish_rollout}), while SMART uses the next, also noised, token as target - effectively treating the noise injection as data augmentation on the training trajectories (c.f. \cref{fig:smart_rollout}).

However, Trajeglish's noisy tokenization does not yield a significant improvement (see Fig. 9 in~\cite{philion2024trajeglish}).
Similarly, while SMART's trajectory perturbation enhances the performance of a zero-shot policy trained on NuPlan~\cite{nuplan} and evaluated on WOSAC, it does not improve performance on WOSAC itself (see Tab. 4 in~\cite{wu2025smart}).
We believe this is because the state distributions generated by their sampling strategies are sub-optimal as they completely ignore what state distribution would be induced by the learned policy, hence likely oversampling irrelevant states and under-sampling states the learned policy would actually encounter. 
By incorporating the learned policy into the sampling strategy, CAT-K generates a state distribution more like the policy's, and is hence better able to reduce the covariate shift between training and inference.
Although sampling from the policy incurs higher data generation costs, these can be effectively mitigated through the two-phase training strategy with BC pre-training and closed-loop fine-tuning.


Lastly, note that CAT-K with $K=|V|$ is equivalent to noise-free BC, since in this case rollouts follow the GT as closely as the available token book allows. 
On the other hand, for $K=1$, CAT-K is equivalent to deterministically rolling out the policy by always choosing the most likely token. 
Consequently, for CAT-K, the hyperparameter $K$ trades off following the policy (for $K=1$) vs. following the GT (for $K=|V|$). In \cref{fig:ade_rollout_gt} we show how various choices of $K$ impact the Average Displacement Error (ADE) between the CAT-K rollouts and the GT.
The intuition behind CAT-K is to find, among all the top-K T-step rollouts (with a total of $K^T$ possibilities), the one with the minimum ADE (minADE) to the GT.
However, since directly solving this optimization problem is difficult, we use CAT-K as a greedy approximation to solve it sequentially.


\section{Experiments}

To show the broad relevance of CAT-K fine-tuning, we evaluate its performance on two different tasks using two different types of policy architecture, namely a \emph{traffic simulation} task using a NTP policy and an \emph{ego-motion planning} task using a policy parameterized as GMM.
We base both tasks on the widely used WOMD~\citep{ettinger2021large} and follow the simulation setup proposed in WOSAC~\citep{montali2024waymo}, namely providing 1 second of history and generating rollouts of 8 seconds length.

\subsection{Traffic Simulation}
\noindent\textbf{Metrics.}\;
In the traffic simulation task, we follow the WOSAC protocol: for each scenario, we generate 32 simulated rollouts \emph{for all agents in the scene}, at 10Hz, and evaluate how well their distribution matches that of the human demonstrations in the data.
This distributional matching is evaluated along three dimensions, namely kinematics (e.g., velocities and accelerations), interactions (e.g., collisions), and map alignment (e.g., off-road). All three are summarized as weighted averages in the ``Realism Meta-Metric" (RMM), the key performance indicator of the WOSAC leaderboard. For more details please see \citep{montali2024waymo}.
We also report the minADE, which is not used in the RMM, but widely applied for motion prediction and policy evaluation.

\begin{table*}[ht!]
\setlength{\tabcolsep}{8pt}
\centering
\scalebox{\tablescaler}{
\begin{tabular}{lrcccccc} 
\toprule
\begin{tabular}{@{}l@{}} \emph{Leaderboard, test split} \\ Method \end{tabular} 
& \begin{tabular}{@{}r@{}} \# model \\ params \end{tabular} 
& \begin{tabular}{@{}c@{}} RMM \\ $\uparrow$ \end{tabular} 
& \begin{tabular}{@{}c@{}} RMM diff. to \\ SMART-large \end{tabular} 
& \begin{tabular}{@{}c@{}} Kinematic \\ metrics $\uparrow$ \end{tabular} 
& \begin{tabular}{@{}c@{}} Interactive \\ metrics $\uparrow$ \end{tabular} 
& \begin{tabular}{@{}c@{}} Map-based \\ metrics $\uparrow$ \end{tabular}
& \begin{tabular}{@{}c@{}} min \\ ADE $\downarrow$ \end{tabular}  \\
\cmidrule(lr){1-2}\cmidrule(lr){3-4}\cmidrule(lr){5-8}
SMART-tiny fine-tuned w. CAT-K (ours) & $7$ M
& $\mathbf{0.7702}$ & $+0.0088$ & $\mathbf{0.4931}$ & $\mathbf{0.8119}$ & $\mathbf{0.8749}$  & $\mathbf{1.3068}$  \\
SMART-large~\cite{wu2025smart} & $102$ M
& $0.7614$ & $+0.0000$ & $0.4786$ & $0.8066$ & $0.8648$  & $1.3728$  \\
KiGRAS~\cite{zhao2024kigras} & $0.7$ M
& $0.7597$  & $-0.0014$  & $0.4691$ & $0.8064$ & $0.8658$  & $1.4384$  \\
SMART-tiny~\cite{wu2025smart} & $7$ M
& $0.7591$ & $-0.0023$ & $0.4759$ & $0.8039$ & $0.8632$  & $1.4062$  \\
FDriver-tiny & $7$ M 
& $0.7584$ & $-0.0030$ & $0.4614$ & $0.8069$ & $0.8658$  & $1.4475$  \\
SMART~\cite{wu2025smart} & $8$ M
& $0.7511$ & $-0.0103$ & $0.4445$ & $0.8050$ & $0.8571$  & $1.5447$  \\
BehaviorGPT~\cite{zhou2024behaviorgpt} & $3$ M
& $0.7473$ & $-0.0141$ & $0.4333$ & $0.7997$ & $0.8593$  & $1.4147$  \\
GUMP~\cite{hu2025solving} & $523$ M
& $0.7431$ & $-0.0183$ & $0.4780$ & $0.7887$ & $0.8359$  & $1.6041$  \\
\bottomrule
\end{tabular}
}
\vspace{-1.5ex}
\caption{\textbf{Results on the WOSAC 2024 leaderboard \cite{wosac2024}}. RMM stands for Realism Meta Metric, the key metric used for ranking. 
Note that on the public leaderboard \cite{wosac2024} our method appears under the name ``SMART-tiny-CLSFT" (Closed-Loop Supervised Fine-Tuning). 
}
\label{table:waymo_test}
\vspace{-3ex}
\end{table*}

\noindent\textbf{Policy.}\;
We use SMART \citep{wu2025smart} as our policy architecture due to its strong performance and open-source implementation.
Specifically, we use the SMART-tiny (7M) model and its open-sourced token vocabularies.
The map polyline token vocabulary has a size of $1024$, with each token representing a polyline consisting of $10$ segments, each $0.5$ meters long.
The agent trajectory token vocabulary has a size of $2048$, with each token representing a $0.5$-second trajectory at $10$Hz.
SMART re-plans at $2$Hz, hence the final output is at $10$Hz when submitted to the WOSAC leaderboard.
However, our experiments show that SMART's open-sourced agent token vocabulary results in lower kinematic metrics.
To address this issue, we reran the K-disk clustering~\cite{philion2024trajeglish} with a larger number of trajectories.
By using our own agent token vocabulary, the performance on the WOSAC leaderboard is significantly improved (c.f. \cref{table:waymo_test}).
Please note that all other experiments in this paper (i.e., except those in \cref{table:waymo_test}) still use SMART's open-sourced token vocabularies.
SMART uses Transformers~\cite{vaswani2017attention} with query-centric representations~\cite{zhou2023query,zhang2024real}, and the distance metric $d(\cdot, \cdot)$ is the average Euclidean distance between the four pairs of corners of two bounding boxes.
We disable SMART's trajectory perturbation as it deteriorates performance on the WOSAC leaderboard (see \citep{wu2025smart} Tab. 4).
We run BC pre-training for $32$ epochs, each taking $1.7$ hour, then continue with closed-loop supervised fine-tuning with CAT-32 for $10$ epochs, each taking $2.6$ hours.
During fine-tuning, the map encoder is frozen to save GPU memory, allowing for a larger training batch size with negligible impact on performance.
For both pre-training and fine-tuning, we use 8$\times$ A100 80GB GPUs with a total batch size of 80.

\subsection{Ego-motion planning}
\noindent\textbf{Metrics.}\;
Unlike traffic simulation, in ego-motion planning only the ego-vehicle is controlled and evaluated.
Moreover, the optimal policy is not the one that perfectly matches the GT trajectory \emph{distribution}, but which minimizes planning metrics, such as collision rate, off-road rate, and ADE.
For completeness, we also report RMM, applied only to ego trajectories, and minADE over 32 sampled rollouts.
With the exception of minADE, and in contrast to the traffic simulation task, the rollouts for evaluation are sampled deterministically, by choosing the most-likely mode at each timestep.

\begin{figure}[t]
  \centering
   \includegraphics[width=0.99\linewidth]{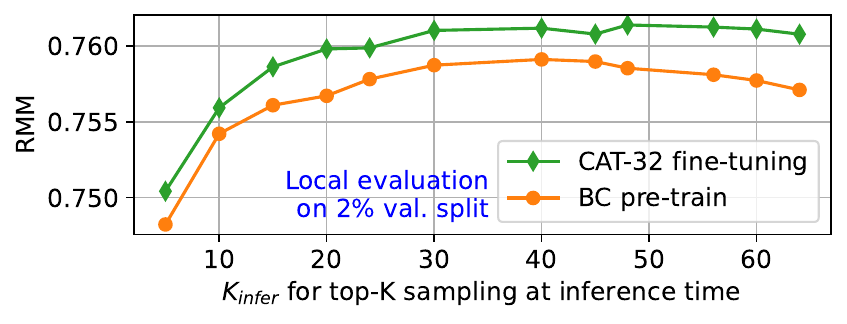}
   \vspace{-2.5ex}
   \caption{Influence of $K_\text{infer}$ for inference-time top-K sampling.}
   \label{fig:influence_val_k}
   \vspace{-2ex}
\end{figure}

\noindent\textbf{Policy.}\;
To show that our fine-tuning approach can be applied to a more general class of policies beyond NTP, we parameterized the output of the ego-policy as GMM.
Specifically, we replace the final layer of the SMART network with two heads, a classification head that predicts the mixture densities and a regression head that predicts the means of the Gaussian distributions with fixed standard deviations (similar to, e.g., \cite{ivanovic2019trajectron}).
Specifically, the GMM predicts 16 modes, each representing a 3-dimensional Gaussian distribution over the changes in x- and y-position and yaw heading.
The final model comprises a total of 3.2M parameters.
In contrast to the SMART traffic simulation policy that uses cross-entropy loss on all agents, the GMM-based ego-policy is trained using negative log-likelihood loss and only predicts actions for the ego-vehicle.
When applying CAT-K rollout or top-K sampling to the GMM, we select the K most-likely modes of the GMM, and within each mode deterministically use the mean of the Gaussian.
Please note that ego-motion planning is a separate experiment from traffic simulation; hence, the ego-policy is not fine-tuned on the traffic simulation policy.
Also note that the ego-policy operates in a mid-to-end manner, taking tracking results and HD maps as inputs, rather than in a fully end-to-end fashion that maps raw sensor observations to actions.

\subsection{Results}

\begin{table*}[ht!]
\setlength{\tabcolsep}{7pt}
\centering
\scalebox{\tablescaler}{
\begin{tabular}{lllllccccc} 
\toprule
\begin{tabular}{@{}l@{}} \emph{Local val. split} \\ Method  \end{tabular} 
& \begin{tabular}{@{}l@{}} Criterion \\ of $\topk{K}$ \end{tabular} 
& \begin{tabular}{@{}l@{}} $K$ for \\ $\topk{K}$  \\  \end{tabular} 
& \begin{tabular}{@{}l@{}} Sampled \\ from \end{tabular} 
& \begin{tabular}{@{}l@{}} Next \\ target \end{tabular} 
& \begin{tabular}{@{}c@{}} RMM \\ $\uparrow$ \end{tabular} 
& \begin{tabular}{@{}c@{}} Kinematic \\ metrics $\uparrow$ \end{tabular} 
& \begin{tabular}{@{}c@{}} Interactive \\ metrics $\uparrow$ \end{tabular} 
& \begin{tabular}{@{}c@{}} Map-based \\ metrics $\uparrow$ \end{tabular}
& \begin{tabular}{@{}c@{}} min \\ ADE $\downarrow$ \end{tabular}
\\ 
\cmidrule(lr){1-1}\cmidrule(lr){2-5}\cmidrule(lr){6-10}
BC pre-training &
- & - & - & GT
& $0.7581$ & $0.4512$ & $0.8076$ & $0.8697$ & $1.3152$ \\
\cmidrule(lr){1-1}\cmidrule(lr){2-5}\cmidrule(lr){6-10}
BC fine-tuning &
- & - & - & GT
& $0.7590$ & $0.4514$ & $0.8096$ & $0.8700$ & $1.3039$ \\
\cmidrule(lr){1-1}\cmidrule(lr){2-5}\cmidrule(lr){6-10}
\multirow{3}{*}{\begin{tabular}{@{}l@{}} Trajeglish's noisy \\ tokenization \end{tabular} }
& neg. dist. & $5^\dagger$ & $\text{neg. dist.}^\dagger$ & $\text{GT}^\dagger$
& $0.7562$ & $0.4469$ & $0.8074$ & $0.8673$ & $1.3459$ \\
& neg. dist. & $5^\dagger$  & uniform & $\text{GT}^\dagger$
& $0.7554$ & $0.4467$ & $0.8069$ & $0.8655$ & $1.3404$ \\
& neg. dist. & 32 & $\text{neg. dist.}^\dagger$ &  $\text{GT}^\dagger$
& $0.7401$ & $0.4174$ & $0.7985$ & $0.8493$ & $1.6669$ \\
\cmidrule(lr){1-1}\cmidrule(lr){2-5}\cmidrule(lr){6-10}
\multirow{3}{*}{\begin{tabular}{@{}l@{}} SMART's trajectory\\ perturbation \end{tabular} }
& neg. dist. & $5^\dagger$ & $\text{uniform}^\dagger$ &  $\text{RO}^\dagger$
& $0.7556$ & $0.4440$ & $0.8082$ & $0.8661$ & $1.3177$ \\
& neg. dist. & $5^\dagger$ & neg. dist. & $\text{RO}^\dagger$
& $0.7560$ & $0.4469$ & $0.8069$ & $0.8673$ & $1.3514$ \\
& neg. dist. & 32 & $\text{uniform}^\dagger$ & $\text{RO}^\dagger$
& $0.7314$ & $0.4158$ & $0.7949$ & $0.8300$ & $1.5380$ \\
\cmidrule(lr){1-10}
Top-5 & prob & 5 & prob & GT 
& $0.6478$ & $0.3313$ & $0.6847$ & $0.7528$ & $1.8802$ \\
Top-5 + distance filter & prob & 5 & prob & GT 
& $0.6860$ & $0.3356$ & $0.7466$ & $0.8083$ & $1.7627$ \\
Top-5 + distance based sampling & prob & 5 & neg. dist. & GT 
& $0.7058$ & $0.3536$ & $0.7579$ & $0.8400$ & $1.5848$ \\
Deterministic rollout & - & - & max-prob & GT 
& $0.6361$ & $0.3291$ & $0.6845$ & $0.7492$ & $1.8695$ \\
\cmidrule(lr){1-1}\cmidrule(lr){2-5}\cmidrule(lr){6-10}
CAT-5 & prob & - & closest & GT 
& $0.7423$ & $0.4251$ & $0.7917$ & $0.8601$ & $1.4677$ \\
CAT-16 & prob & - & closest & GT 
& $0.7604$ & $\textbf{0.4592}$ & $0.8082$ & $0.8709$ & $1.3372$ \\
\textbf{CAT-32 (leaderboard)} & prob & - & closest & GT 
& $0.7616$ & $0.4583$ & $\textbf{0.8105}$ & $0.8720$ & $1.3105$ \\
CAT-40 & prob & - & closest & GT 
& $\textbf{0.7617}$ & $0.4567$ & $0.8101$ & $\textbf{0.8738}$ & $\textbf{1.2998}$ \\
CAT-64 & prob & - & closest & GT 
& $0.7602$ & $0.4552$ & $0.8098$ & $0.8707$ & $1.3028$ \\
\bottomrule
\end{tabular}
}
\vspace{-2ex}
\caption{
\textbf{Ablation study on WOSAC 2\% validation split.}
We compare different ways to fine-tune the same base mode (BC pre-training). 
"Sampled from" indicates how the action is sampled during fine-tuning, either based on the distance to the GT (``neg. dist", ``uniform", ``closest") or based on the model outputs (``prob", ``max-prob").
Here dist. is the abbreviation of distance.
RO stands for rollout, i.e., the next target action is computed based on the rollout, not the GT state.
$\dagger$ indicates original hyperparameter choices of baseline algorithms.
}
\label{table:ablation}
\vspace{-4ex}
\end{table*}

\subsubsection{WOSAC leaderboard for traffic simulation}

In \cref{table:waymo_test} we compare our approach with other traffic simulation policies on the WOSAC leaderboard.
Notably, all the top-ranking methods on the leaderboard are NTP policies trained via BC.
Our SMART-tiny model using CAT-K fine-tuning outperforms the previous SoTA, SMART-large with 102M parameters, by a significant margin of $+0.0088$; and improves on SMART-tiny by $+0.0111$.
As the first to perform closed-loop fine-tuning on the leaderboard, our approach improves all metrics and sets a new SoTA. 

For fine-tuning, we chose CAT-32 after preliminary hyperparameter explorations, a choice that was later confirmed to perform well in our ablation studies (see \cref{table:ablation}), though CAT-K improved performance for a wide range of values of $K$.
We also found that choosing a sufficiently high $K_\text{infer}$ for top-K sampling during inference is important for a high RMM~\cite{hu2025solving}, and we chose $K_\text{infer}=48$ for our leaderboard submission based on local validation results (see \cref{fig:influence_val_k}).
In all of our experiments, we fixed the inference time sampling temperature to $1.0$ and did not use top-p (nucleus) sampling~\cite{Holtzman2020The}, though we expect that tuning these hyperparameters can lead to further performance gains.

To ensure that the observed performance gains with CAT-K are not due to improved hyperparameter tuning, we also conduct a large-scale hyperparameter grid search for the baseline method SMART-tiny. This includes training for more epochs, adjusting the learning rate and learning rate scheduler, experimenting with various data augmentation and data pre-processing strategies, and using various values of $K_\text{infer}$ for the inference-time top-K sampling.
This allowed us to push the performance of the BC baseline to $0.7671$ RMM on the test split (not shown in \cref{table:waymo_test}), but still falling significantly short of the performance of our CAT-K fine-tuning method ($0.7702$ RMM).
Note that even comparing against this improved baseline, our method improves the performance more ($+0.0031$) than scaling up the model size by a factor of 14 to 102M parameters ($+0.0023$).

Qualitatively, our method can generate diverse and realistic behavior over long periods of time.
As shown in \cref{fig:qualitative_wosac}, our method can handle the subtle interactions between traffic participants in a dense parking lot, which is arguably a more challenging scenario than intersections and highways for traffic simulation.
Moreover, the behavior remains realistic at the end of the required simulation time of 8 seconds. Additional examples are in the supplementary videos.

\begin{figure}[t!]
  \centering
   \includegraphics[width=0.99\linewidth]{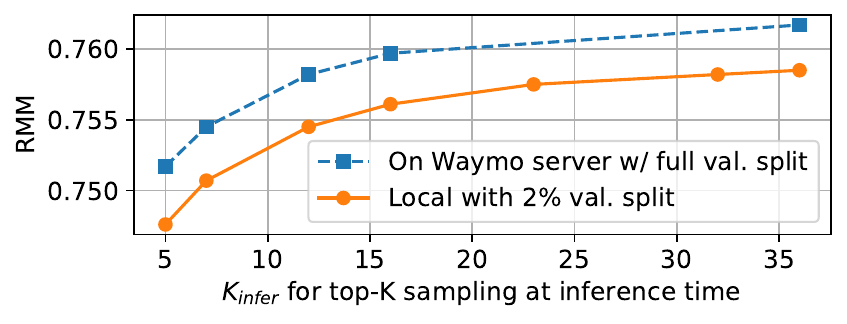}
   \vspace{-3ex}
   \caption{On server vs. local evaluation of SMART-tiny.}
   \label{fig:local_vs_server}
   \vspace{-2ex}
\end{figure}

\subsubsection{Ablation studies on WOSAC}

In \cref{table:ablation} we provide a thorough ablation study.
Due to the high cost of evaluation we use 2\% of the validation split (880 out of 44097 scenarios). To verify the fidelity of this evaluation setting, we compare results with those on the full validation set in \cref{fig:local_vs_server}. We observe consistent differences indicating that our evaluation setting is reasonable. 
We begin with a SMART-tiny model trained with BC for 32 epochs (BC pre-training, row 1). All other models fine-tune the BC pre-training model for 5 epochs. During inference, we use top-K sampling with $K_\text{infer}=40$ for all methods (the best $K_\text{infer}$ for SMART-tiny according to \cref{fig:influence_val_k}).

In \cref{table:ablation}, CAT-K fine-tuning is the only method that significantly outperforms the BC pre-training model. Further fine-tuning with BC (row 2), using Trajeglish's noisy tokenization (rows 3-5), or SMART's trajectory perturbation (rows 6-8) remain on par with the original model or even reduce its performance.
Closed-loop fine-tuning with top-K rollouts instead of CAT-K rollouts during training significantly reduces performance (rows 9-12), even when rollouts close to the GT are selected for training by either filtering them based on the distance to the GT or sampling among them from a distance-dependent distribution. 
The observation that data augmentation and fine-tuning with top-K sampling cannot improve the RMM is consistent with prior works~\cite{philion2024trajeglish,wu2025smart,lu2023imitation,peng2025improving} and the WOSAC leaderboard.
Ablations on different values of $K$ for top-K sampling and other hyperparameters are provided in the appendix.

\begin{figure}[t!]
  \centering
   \includegraphics[width=0.91\linewidth]{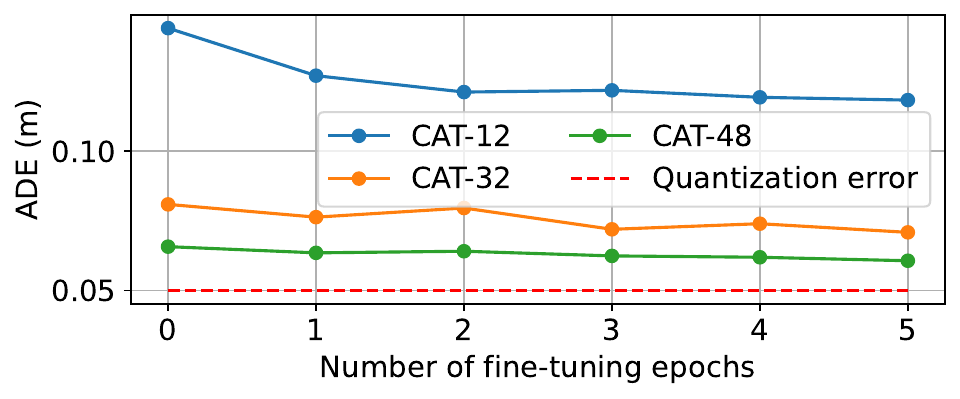}
   \vspace{-2.9ex}
   \caption{ADE between CAT-K rollouts and GT trajectories.}
   \label{fig:ade_rollout_gt}
   \vspace{-2ex}
\end{figure}

\begin{figure*}
  \centering

  \begin{subfigure}{0.195\linewidth}
    \includegraphics[width=0.99\linewidth]{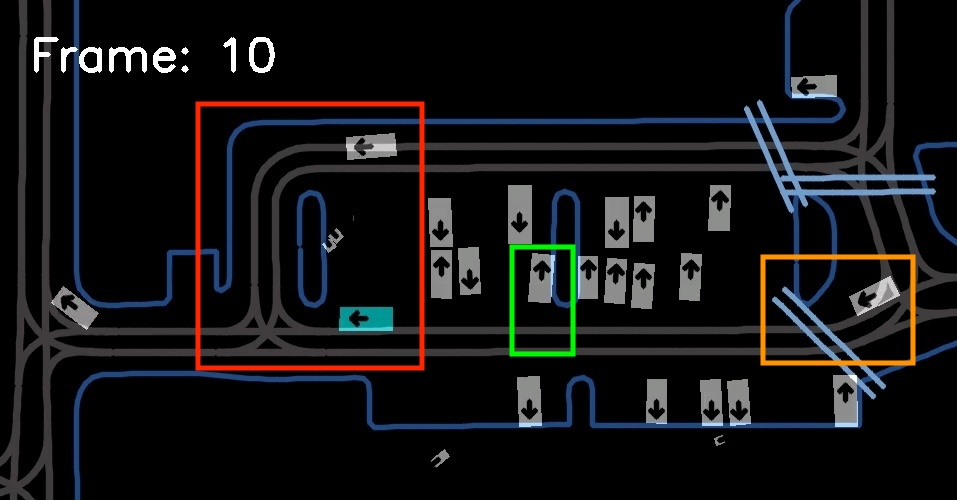}
  \end{subfigure}
  \hfill
  \begin{subfigure}{0.195\linewidth}
    \includegraphics[width=0.99\linewidth]{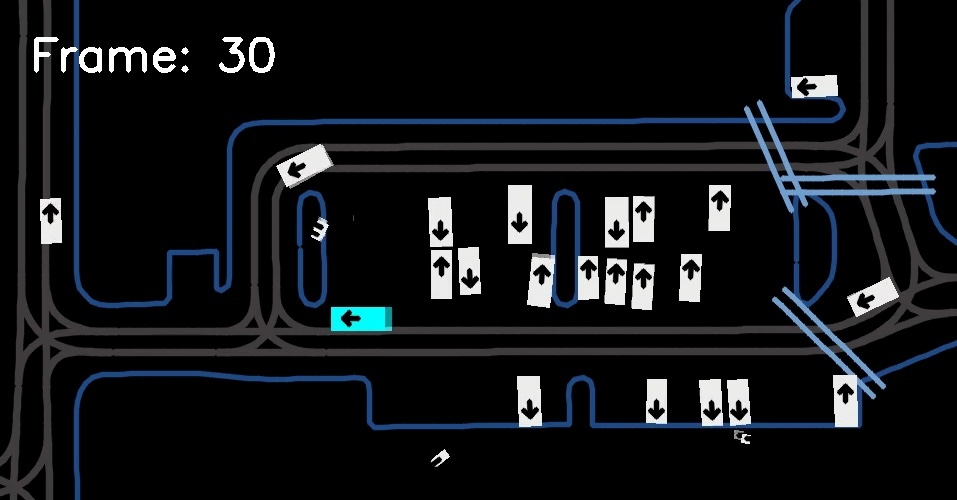}
  \end{subfigure}
  \hfill
  \begin{subfigure}{0.195\linewidth}
    \includegraphics[width=0.99\linewidth]{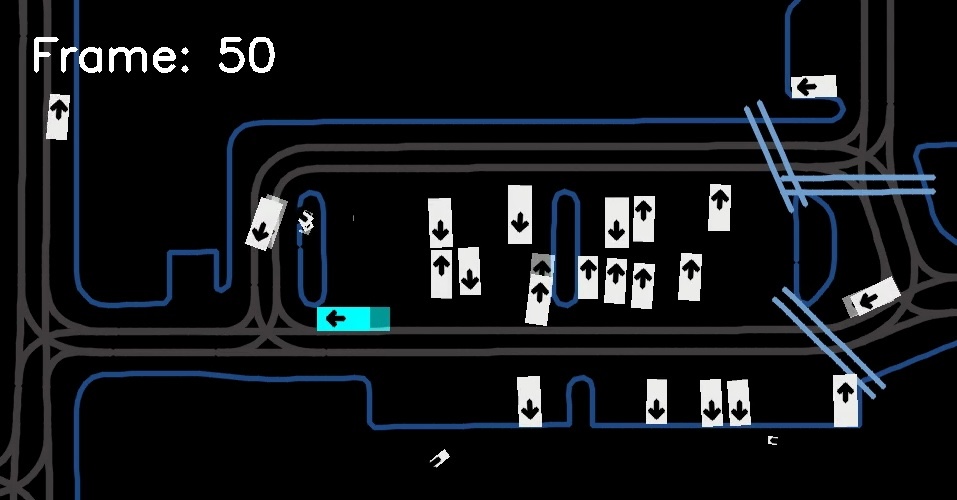}
  \end{subfigure}
  \hfill
  \begin{subfigure}{0.195\linewidth}
    \includegraphics[width=0.99\linewidth]{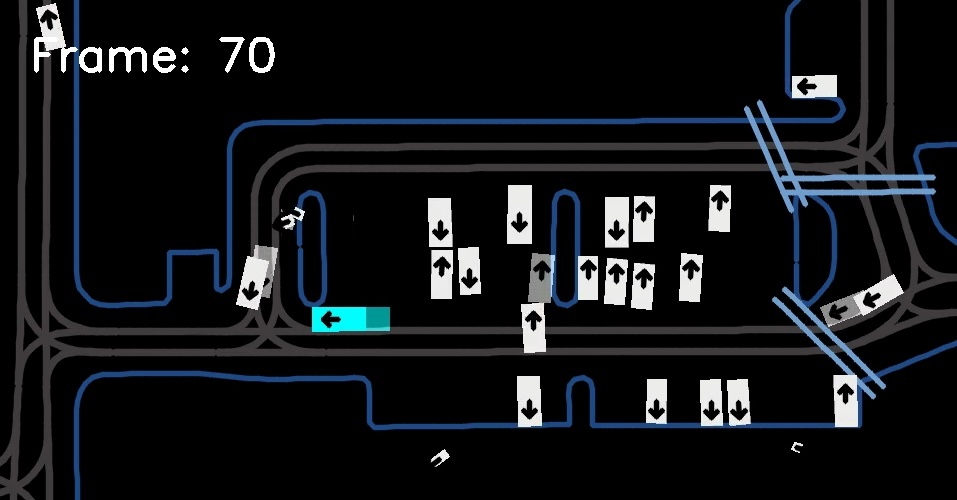}
  \end{subfigure}
  \hfill
  \begin{subfigure}{0.195\linewidth}
    \includegraphics[width=0.99\linewidth]{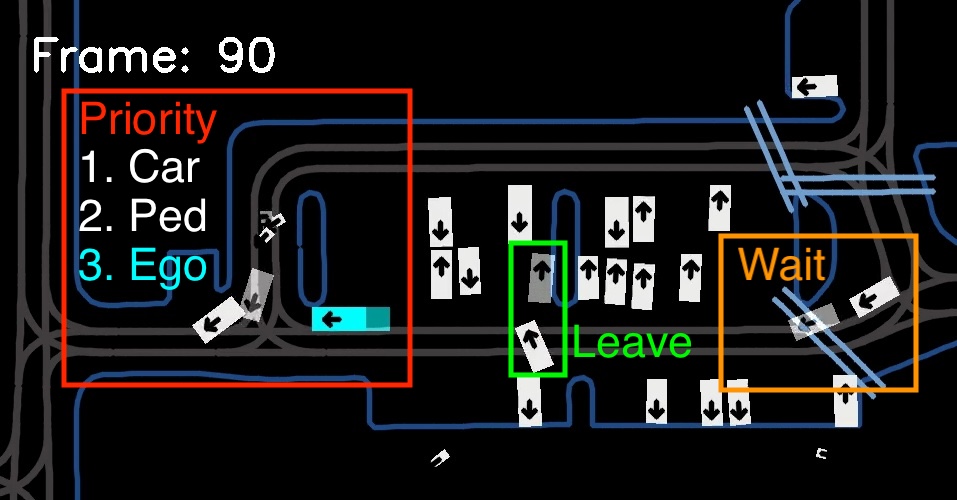}
  \end{subfigure}
  \hfill
  
  \begin{subfigure}{0.195\linewidth}
    \includegraphics[width=0.99\linewidth]{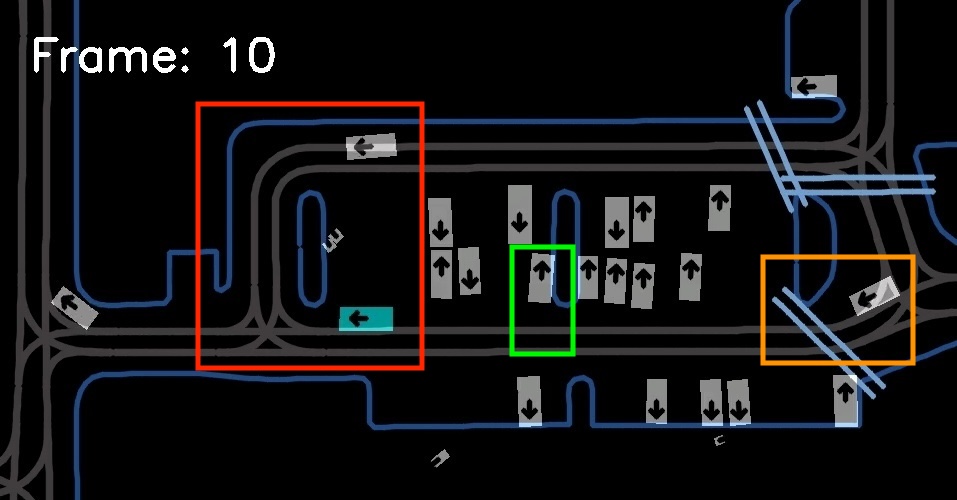}
  \end{subfigure}
  \hfill
  \begin{subfigure}{0.195\linewidth}
    \includegraphics[width=0.99\linewidth]{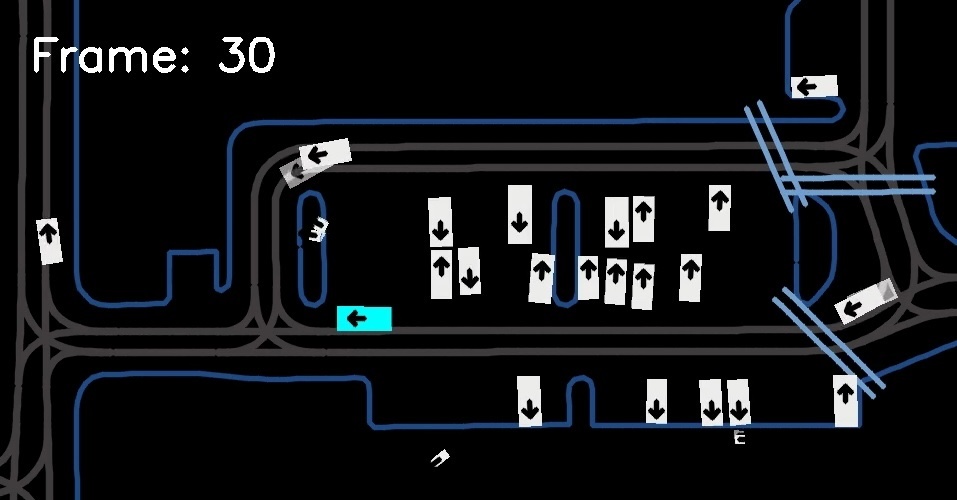}
  \end{subfigure}
  \hfill
  \begin{subfigure}{0.195\linewidth}
    \includegraphics[width=0.99\linewidth]{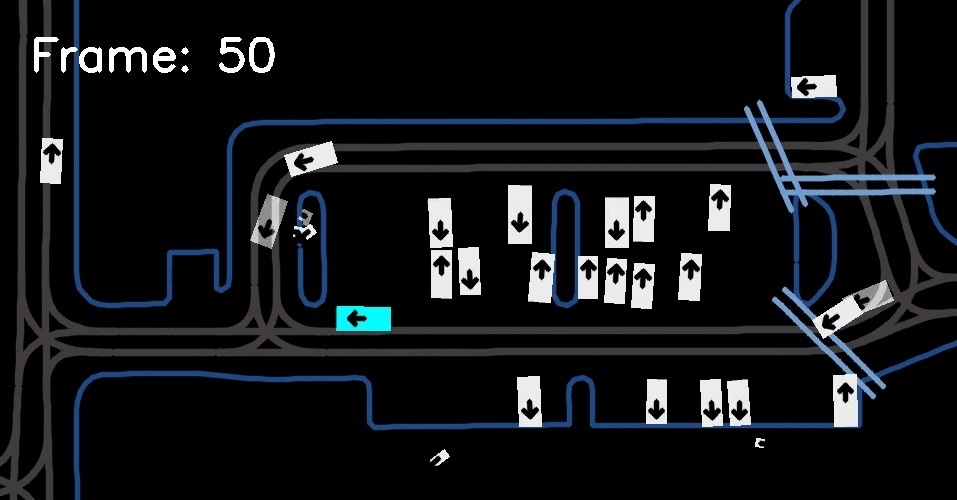}
  \end{subfigure}
  \hfill
  \begin{subfigure}{0.195\linewidth}
    \includegraphics[width=0.99\linewidth]{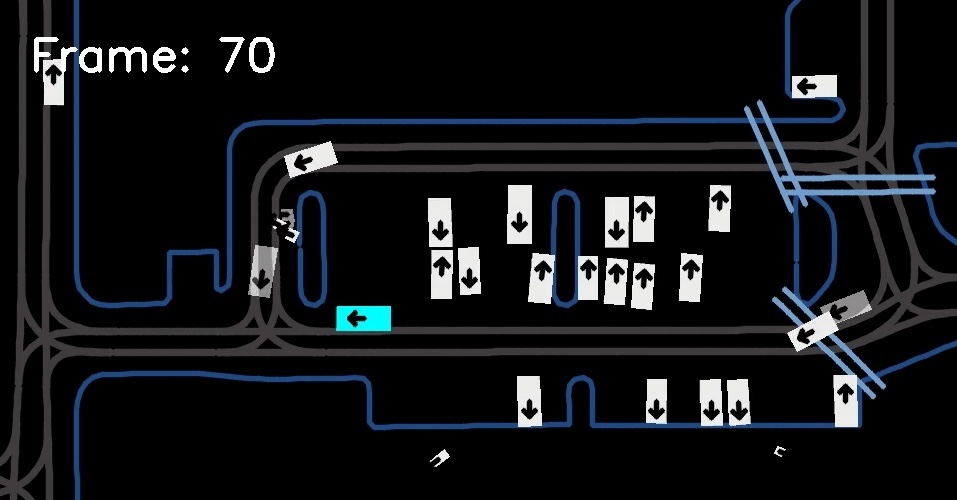}
  \end{subfigure}
  \hfill
  \begin{subfigure}{0.195\linewidth}
    \includegraphics[width=0.99\linewidth]{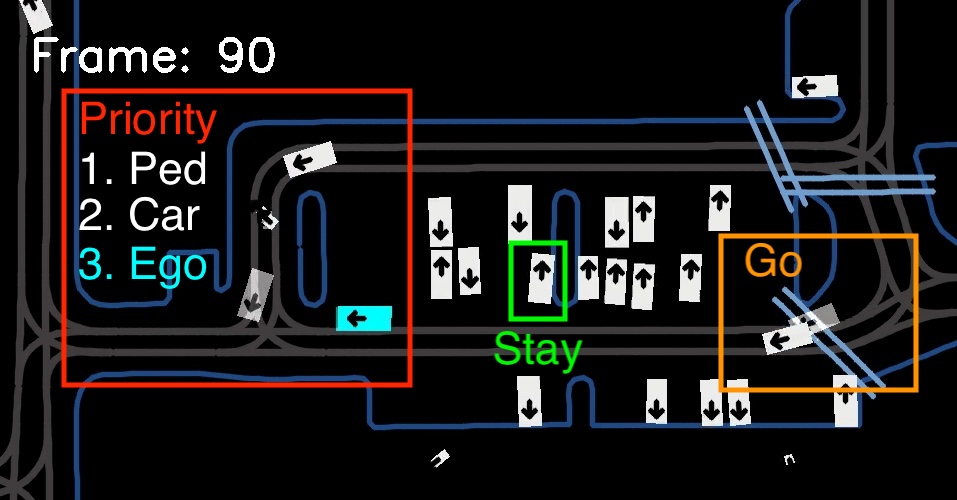}
  \end{subfigure}
   \hfill

  \begin{subfigure}{0.195\linewidth}
    \includegraphics[width=0.99\linewidth]{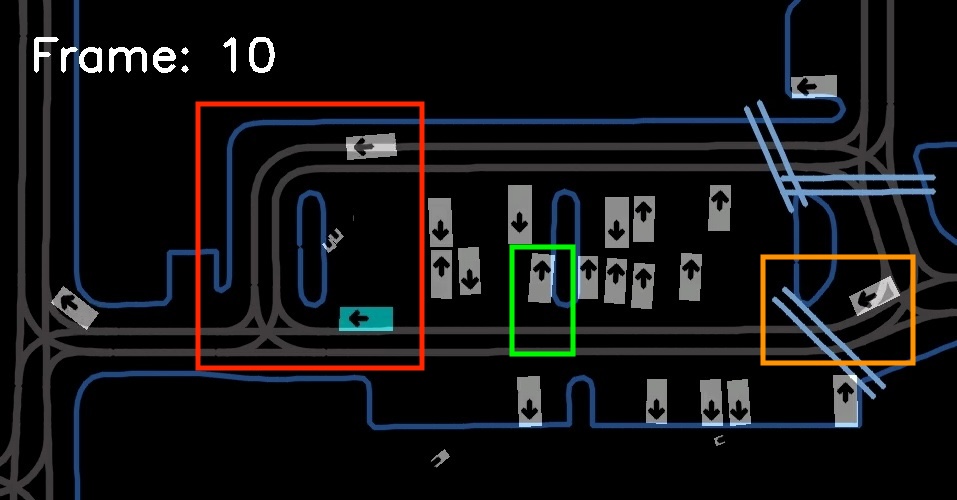}
  \end{subfigure}
  \hfill
  \begin{subfigure}{0.195\linewidth}
    \includegraphics[width=0.99\linewidth]{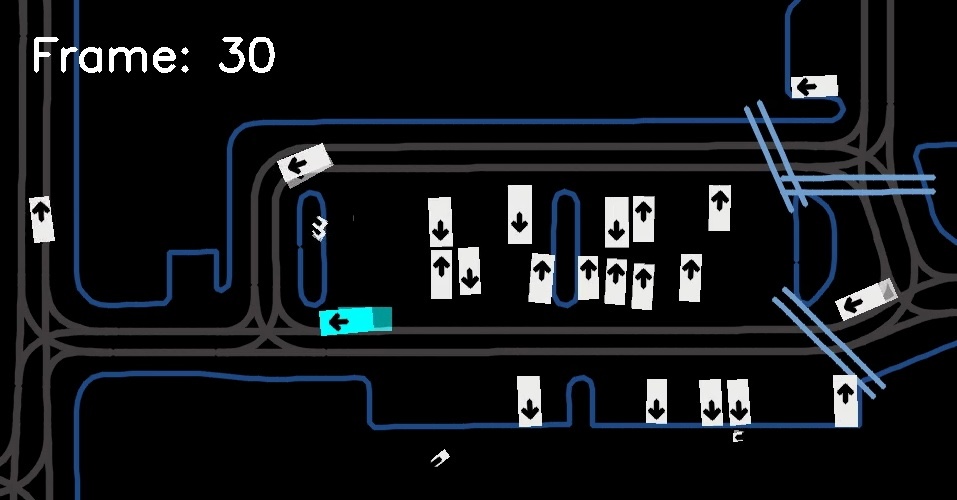}
  \end{subfigure}
  \hfill
  \begin{subfigure}{0.195\linewidth}
    \includegraphics[width=0.99\linewidth]{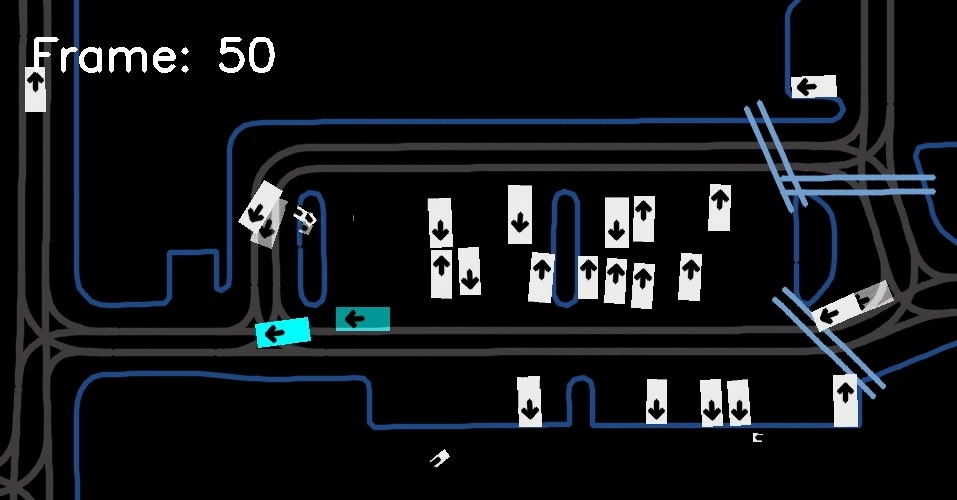}
  \end{subfigure}
  \hfill
  \begin{subfigure}{0.195\linewidth}
    \includegraphics[width=0.99\linewidth]{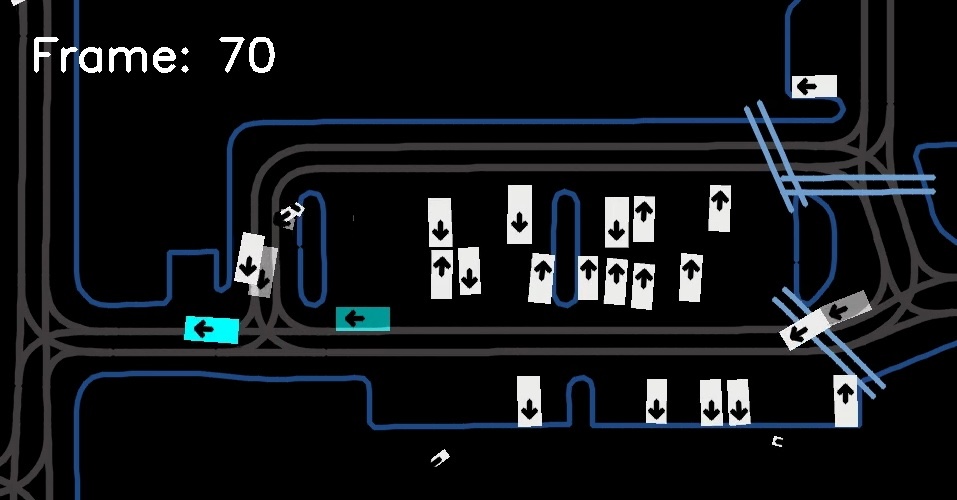}
  \end{subfigure}
  \hfill
  \begin{subfigure}{0.195\linewidth}
    \includegraphics[width=0.99\linewidth]{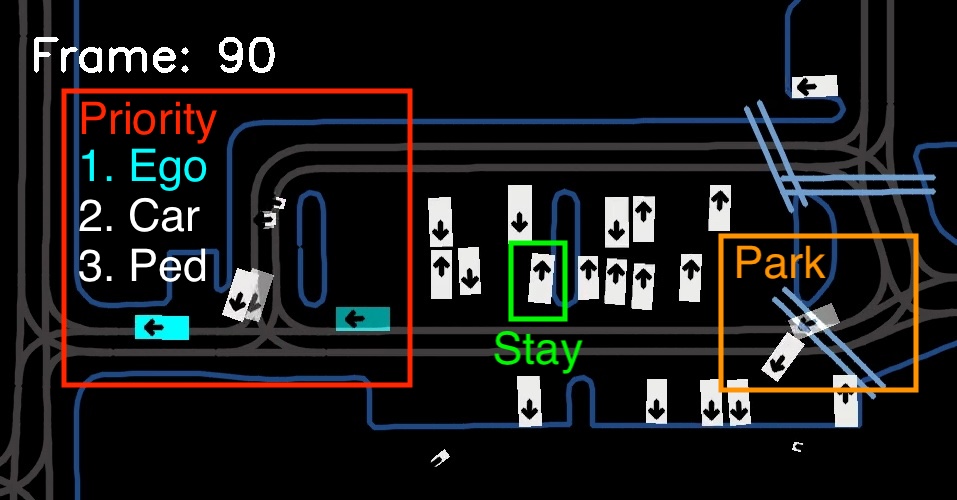}
  \end{subfigure}
  \hfill
  \vspace{-1.5ex}
  \caption{
  \textbf{Simulation results on WOSAC.}
  Our fine-tuned policy generates interesting and diverse behaviors rarely seen in prior works.
  Each row represents a different rollout of our model in the same scene. The transparent boxes show the GT agents in the dataset, while the solid boxes show the agents generated by our model. We highlight the agents within the red, green, and orange rectangles, across time steps and rollouts. The red rectangle shows different interactive negotiations emerging between a pedestrian and two vehicles. The green rectangle shows an initially parked vehicle, that leaves (row 1) or stays parked (row 2, 3). The orange rectangle shows a vehicle waiting (row 1) in front of a speed bump (visualized as two light-blue lines), proceeding (row 2), or entering a parking space (row 3).
  }
  \label{fig:qualitative_wosac}
\vspace{-1ex}
\end{figure*}

\begin{table*}
\setlength{\tabcolsep}{10pt}
\centering
\scalebox{\tablescaler}{
\begin{tabular}{lccccc} 
\toprule
\begin{tabular}{@{}l@{}} Method (\emph{Local val. split}) \end{tabular} 
& \begin{tabular}{@{}c@{}} Collision rate $\downarrow$ \end{tabular} 
& \begin{tabular}{@{}c@{}} Off-road rate $\downarrow$ \end{tabular} 
& \begin{tabular}{@{}c@{}} RMM $\uparrow$ \end{tabular} 
& \begin{tabular}{@{}c@{}} ADE $\downarrow$ \end{tabular} 
& \begin{tabular}{@{}c@{}} minADE$^{32}$ $\downarrow$ \end{tabular} 
\\
\cmidrule(lr){1-1}\cmidrule(lr){2-5}\cmidrule(lr){6-6}
BC pre-training &
$0.0568$ & $0.0053$ & $0.8108$ & $1.3623$  & $1.3537$ \\
\cmidrule(lr){1-1}\cmidrule(lr){2-5}\cmidrule(lr){6-6}
BC fine-tuning &
$0.0599$ & $0.0058$ & $0.8105$ & $1.3520$ & $1.3509$ \\
Deterministic rollout &
$0.0433$ & $0.0138$ & $0.8081$ & $\mathbf{1.1799}$ & $0.7962$ \\
CAT-3 &
$\mathbf{0.0422}$ & $\mathbf{0.0035}$ & $\mathbf{0.8169}$ & $1.3096$ & $\mathbf{0.6912}$ \\
\bottomrule
\end{tabular}
}
\vspace{-1.5ex}
\caption{\textbf{Performance of ego policies on WOSAC with local evaluation on 2\% validation split.}
All models are fine-tuned for 5 epochs based on the BC pre-training model, which is trained for 32 epochs.
We use deterministic rollout during inference and compute all metrics, except for the minADE$^{32}$.
For minADE$^{32}$, we generate 32 rollouts by using top-3 sampling with a temperature of $1.0$ to first sample the categorical distribution over the mixtures, then selecting the mean of the sampled Gaussian mixture.
}
\label{table:ego_policy}
\vspace{-1ex}
\end{table*}

Next, we ablate the value of $K$ used for CAT-K rollout during fine-tuning (rows 13-17). Results indicate that the performance improvement is robust to the choice of $K$ after a reasonable minimum value. 
As discussed in \cref{sec:comparison_prev_methods}, the hyperparameter $K$ in CAT-K determines how closely the policy follows the GT trajectory, in a way that is more robust than a distance based threshold, for which the optimal value varies strongly based on the situation (vehicle speed, proximity to other cars, etc.).
To give more insight into how $K$ impacts rollouts, in \cref{fig:ade_rollout_gt}, we inspect the average ADE between rollouts and GT trajectories over training epochs, for different $K$ values.
As expected, as $K\to|V|$, the ADE decreases towards the level of quantization error, induced by tokenization with a finite vocabulary size (dashed line). 
With more fine-tuning epochs the average ADE slightly reduces, highlighting how CAT-K fine-tuning improves the policy to follow all behavior modes more closely.

\subsubsection{Fine-tuning a GMM-based ego-policy}

Besides NTP traffic simulation polices, CAT-K fine-tuning also improves the performance of a GMM-based ego-policy with continuous action space.
In \cref{table:ego_policy} we compare our approach with fine-tuning using deterministic rollouts, as well as with continued BC.
Our CAT-3 fine-tuning improves all metrics except for ADE, where deterministic rollout performs better.
This is expected, as deterministic rollout aligns all modes towards the GT, resulting in mode averaging.
While this reduces ADE, it negatively impacts other metrics.
Additionally, CAT-K fine-tuning helps overcome the limitation of GMM trained with negative log-likelihood loss, which struggles to capture multimodality as effectively as NTP policies trained with cross-entropy loss.
This is supported by the significant drop in minADE after fine-tuning.

\section{Conclusion}
In this paper, we introduce CAT-K rollouts, a closed-loop supervised fine-tuning technique for IL problems with highly multimodal demonstrations, such as traffic simulation.
The CAT-K rollout approximately finds the rollout closest to the GT among likely rollouts of a policy, ensuring adherence to the policy while maintaining GT as a reliable reference for supervised learning. As the first method using closed-loop fine-tuning, it achieves the top spot on the WOSAC leaderboard.

In the future, we aim to incorporate modern sampling techniques into CAT-K rollouts, such as top-p sampling.
Furthermore, instead of always selecting the closest token, we could loosen the ``winner-takes-all'' approach and sample from a group of nearby tokens.
We also plan to explore CAT-K fine-tuning for a broader range of policy classes, such as variational autoencoders and diffusion models.
Additionally, we want to apply it to other multimodal IL tasks, including end-to-end driving, motion generation for animation, and robot navigation and manipulation.
Importantly, our results show that closed-loop supervised fine-tuning is a promising area of future research for policies trained in open-loop, such as the widely used NTP policies.


\clearpage
{
    \small
    \bibliographystyle{ieeenat_fullname}
    \bibliography{main}
}
\appendix
\clearpage

\section{Supplementary videos}

We provide the following videos as part of the supplementary material.
All videos are carefully edited and thoroughly annotated, offering additional qualitative results to support our paper.
Please note that all videos are without sound.
\begin{itemize}
    \item \texttt{tldr\_highlights.mp4}: The most interesting behaviors generated by our model. If you are short on time, we recommend watching this video.
    \item \texttt{parking\_lot.mp4}: The busy parking lot scenario shown in the main paper.
    \item \texttt{ped\_cyc.mp4}: Interesting behaviors for pedestrians and cyclists.
    \item \texttt{lane\_changing.mp4}: Lane-changing scenario involving interactions between multiple agents.
    \item \texttt{bc\_compounding\_error.mp4}: This video provides a real-world example of covariate shift, demonstrating how the BC policy suffers from compounding errors.
    \item \texttt{more\_behaviors.mp4}: If you have time, enjoy additional interesting behaviors generated by our model. These include exiting parking lots, making U-turns, stopping at stop signs, obeying traffic lights, near-accident scenarios, and more.
\end{itemize}
Additional videos are available on the homepage of this paper\footnote{\url{https://zhejz.github.io/catk}}.

\begin{algorithm}[t]
    \caption{BC pre-training and CAT-K fine-tuning}
    \label{alg:bc_and_catk}
\begin{algorithmic}[1]
    \State \textbf{Input}: Policy $\pi_\theta$, action token vocabulary $V$, dataset $\mathcal{D}$
    \State \multiline{Pre-train $\pi_\theta(\mathbf{c}_t\mid \hat{\mathbf{h}}_t, \mathcal{M})$ with BC until convergence}
    \Repeat \Comment{BC pre-training}
        \State \multiline{Sample a traffic scenario $\{\hat{\mathbf{s}}_{0:T},\mathcal{M}\}$.}
        \State \multiline{Init rollout state $\mathbf{s}_0=\hat{\mathbf{s}}_0$.\\ \Comment{Sequential tokenization following Trajeglish}}
        \For{$t$ in $[0, \dots, T-1]$}
            \State \multiline{Tokenization for each agent $i\in\{1,\dots,N\}$\\
            $c^i_t = \argmin\limits_{c\in \{ 1,\dots,|V| \} } d\left( f(s^i_t, x_c) - \hat{s}^i_{t+1} \right)$.
            }
            \State \multiline{Save $\mathbf{c}_{t}$ as the GT labels $\hat{\mathbf{c}}_{t}$.}
            \State \multiline{Get next rollout state $s^i_{t+1}$.  (Eq.~\ref{eq:catk_update_state})}
        \EndFor
        \State \multiline{Batched forward pass with causal masking \\
        $\pi_\theta(\mathbf{c}_{1:T} \mid \hat{\mathbf{c}}_{0:T-1}, \mathcal{M})$.}
        \State \multiline{Update $\theta$ by minimizing the cross entropy loss Eq.~\ref{eq:cross_entropy} with GT labels $\hat{\mathbf{c}}$.}
    \Until{convergence}
    \Repeat \Comment{Closed-loop supervised fine-tuning}
        \State \multiline{Sample a traffic scenario $\{\hat{\mathbf{s}}_{0:T},\mathcal{M}\}$}
        \State \multiline{Init rollout state $\mathbf{s}_0=\hat{\mathbf{s}}_0$ \Comment{CAT-K Rollout}}
        \For{$t$ in $[0, \dots, T-1]$} \Comment{$T$ steps}
            \For{$i$ in $[1, \dots, N]$} \Comment{$N$ agents}
                \State \multiline{One step forward pass policy $\pi$ with previous rollout states.}
                \State \multiline{Get action index for rollout $c^i_t$. (Eq.~\ref{eq:catk})}
                \State \multiline{Get next rollout state $s^i_{t+1}$.  (Eq.~\ref{eq:catk_update_state})}
                \State \multiline{Compute target $\hat{c}^i_{t}$. (Eq.~\ref{eq:tokenize}) }
                \State \multiline{Save forward-pass output logits of this step for later training.}
            \EndFor
        \EndFor
        \State \multiline{Update $\theta$ by minimizing $\mathcal{L}_\theta(\mathbf{s}_{0:T}, \hat{\mathbf{c}}_{1:T}, \mathcal{M})$. (Eq.~\ref{eq:cross_entropy}) }
    \Until{convergence}
\end{algorithmic}
\end{algorithm}

\section{Full algorithm}
To complement the algorithm in the main paper, we provide the detailed and complete algorithm for BC pre-training, followed by CAT-K fine-tuning, in \cref{alg:bc_and_catk}.
Our BC pre-training follows Trajeglish and SMART, which are identical when no data augmentation is applied.

\begin{table*}[t]
\setlength{\tabcolsep}{2.6pt}
\centering
\scalebox{\tablescaler}{
\begin{tabular}{lccccccccccc} 
\toprule
\begin{tabular}{@{}l@{}} \emph{Leaderboard, test split} \\ Method \end{tabular} 
& \begin{tabular}{@{}c@{}} Realism \\ meta \\ metric$\uparrow$ \end{tabular} 
& \begin{tabular}{@{}c@{}} Linear \\ speed \\ likeli.$\uparrow$ \end{tabular} 
& \begin{tabular}{@{}c@{}} Linear \\ acc. \\ likeli.$\uparrow$ \end{tabular} 
& \begin{tabular}{@{}c@{}} Angular \\ speed \\ likeli.$\uparrow$ \end{tabular} 
& \begin{tabular}{@{}c@{}} Angular \\ acc. \\ likeli.$\uparrow$ \end{tabular} 
& \begin{tabular}{@{}c@{}} Distance to \\ nearest \\ object likeli.$\uparrow$ \end{tabular} 
& \begin{tabular}{@{}c@{}} Collision \\ likeli. \\ $\uparrow$ \end{tabular} 
& \begin{tabular}{@{}c@{}} Time to \\ collision \\ likeli.$\uparrow$ \end{tabular} 
& \begin{tabular}{@{}c@{}} Distance to \\ road edge \\ likeli.$\uparrow$ \end{tabular} 
& \begin{tabular}{@{}c@{}} Offroad \\  likeli. \\ $\uparrow$ \end{tabular}
& \begin{tabular}{@{}c@{}} min \\ ADE \\ $\downarrow$ \end{tabular}  \\
\cmidrule(lr){1-1}\cmidrule(lr){2-12}
SMART-tiny-CLSFT (ours) 
& $\mathbf{0.7702}$ & $\mathbf{0.3868}$ & $0.4066$ & $\mathbf{0.5201}$ & $\mathbf{0.6589}$ & $\mathbf{0.3923}$  & $\mathbf{0.9702}$ & $0.8356$ & $\mathbf{0.6814}$ & $\mathbf{0.9523}$ & $1.3068$ \\
UniMM~\cite{lin2025revisitmixturemodelsmultiagent}
& $0.7683$ & $0.3836$ & $\mathbf{0.4159}$ & $0.5168$ & $0.6491$ & $0.3911$  & $0.9679$ & $0.8347$ & $0.6791$ & $0.9506$ & $\mathbf{1.2947}$ \\
SMART-large~\cite{wu2025smart} 
& $0.7614$ & $0.3786$ & $0.4134$ & $0.4952$ & $0.6270$ & $0.3872$  & $0.9632$ & $0.8346$ & $0.6761$ & $0.9403$ & $1.3728$ \\
KiGRAS~\cite{zhao2024kigras}
& $0.7597$ & $0.3704$ & $0.3784$ & $0.4962$ & $0.6314$ & $0.3867$  & $0.9619$ & $\mathbf{0.8373}$ & $0.6723$ & $0.9431$ & $1.4383$ \\
SMART-tiny~\cite{wu2025smart} 
& $0.7591$ & $0.3733$ & $0.4082$ & $0.4945$ & $0.6277$ & $0.3835$  & $0.9601$ & $0.8338$ & $0.6709$ & $0.9401$ & $1.4062$ \\
FDriver-tiny
& $0.7584$ & $0.3661$ & $0.3669$ & $0.4876$ & $0.6248$ & $0.3840$  & $0.9641$ & $0.8366$ & $0.6688$ & $0.9446$ & $1.4475$ \\
SMART~\cite{wu2025smart} 
& $0.7511$ & $0.3646$ & $0.4057$ & $0.4231$ & $0.5845$ & $0.3769$  & $0.9655$ & $0.8318$ & $0.6590$ & $0.9363$ & $1.5447$ \\
BehaviorGPT~\cite{zhou2024behaviorgpt} 
& $0.7473$ & $0.3615$ & $0.3365$ & $0.4806$ & $0.5544$ & $0.3834$  & $0.9537$ & $0.8308$ & $0.6702$ & $0.9349$ & $1.4147$ \\
GUMP~\cite{hu2025solving} 
& $0.7431$ & $0.3569$ & $0.4111$ & $0.5089$ & $0.6353$ & $0.3707$  & $0.9403$ & $0.8276$ & $0.6686$ & $0.9028$ & $1.6031$ \\
\cmidrule(lr){1-1}\cmidrule(lr){2-12}
\begin{tabular}{@{}l@{}} SMART-tiny (we reproduced) \\not on the public leaderboard \end{tabular}
& $0.7671$ & $0.3781$ & $0.4026$ & $0.5183$ & $0.6571$ & $0.3899$  & $0.9653$ & $0.8346$ & $0.6788$ & $0.9507$ & $1.3587$ \\
\bottomrule
\end{tabular}
}
\caption{\textbf{Results on the WOSAC 2024 leaderboard \cite{wosac2024}} accessed on March 14, 2025.
Realism Meta Metric is the key metric used for ranking.
All other metrics contribute to the realism meta metric, except for the minADE, which has no effect on the ranking.
Note that on the public leaderboard \cite{wosac2024} our method appears under the name ``SMART-tiny-CLSFT" (Closed-Loop Supervised Fine-Tuning), and our reproduced SMART-tiny is not published to the public leaderboard.
Here likeli. is the abbreviation of likelihood, and acc. stands for acceleration.
}
\label{table:waymo_test_full}
\end{table*}

\section{Implementation details}

Our implementation is based on the open-source repository of SMART\footnote{\url{https://github.com/rainmaker22/SMART}}.
We made the following changes, as we believe they may improve performance:
\begin{itemize}
    \item Preprocessing agent trajectories using linear interpolation.
    \item Adding additional HD map elements, such as speed bumps.
    \item Setting the learning rate decay to 1\% instead of to 0\%.
    \item Resolving duplicated tokens in the action token vocabulary.
    \item Removing data augmentation applied to the tokenization of map polylines and agent trajectories, as it only improves performance for zero-shot transfer from NuPlan to WOSAC but significantly decreases performance when the model is both trained and validated on WOSAC~\cite{wu2025smart}. 
\end{itemize}

Apart from these changes, we use the same model architecture, hyperparameters, and other settings as provided in the open-source repository.
While SMART-tiny was originally trained on 32 NVIDIA TESLA V100 GPUs for 23 hours, we use 8 NVIDIA A100 GPUs for all our experiments.
Our reproduced SMART-tiny model is trained for 32 hours (32 epochs) with BC.
We finetune this BC baseline model with CAT-K rollout for 25 hours (10 epochs) to obtain our final model, SMART-tiny-CLSFT, which is submitted to the leaderboard.
Performing inference and generating the submission file for the validation split (44,097 scenarios) together requires 3 hours, the same as for the test split (44,920 scenarios).

\section{Additional experiment results}

\subsection{WOSAC leaderboard}

In \cref{table:waymo_test_full} we provide the results of all metrics for leading entries on the WOSAC leaderboard\footnote{\url{https://waymo.com/open/challenges/2024/sim-agents/}}, accessed before the camera-ready deadline of CVPR 2025 (March 14, 2025).
We also provide the results of our reproduced SMART-tiny, trained via BC and used as the starting point for our fine-tuning experiments.
Our method achieves the best performance across nearly all metrics.
Notably, a concurrent work, UniMM~\cite{lin2025revisitmixturemodelsmultiagent} (previously called MM-GPT), has recently made multiple submissions and achieved a high ranking on the leaderboard.
Nevertheless, our method still outperforms the best UniMM model in the majority of the metrics.


\begin{table*}[t]
\setlength{\tabcolsep}{5pt}
\centering
\scalebox{\tablescaler}{
\begin{tabular}{lllllccccc} 
\toprule
\begin{tabular}{@{}l@{}} \emph{Local val. split} \\ Method  \end{tabular} 
& \begin{tabular}{@{}l@{}} Criterion \\ of $\topk{K}$ \end{tabular} 
& \begin{tabular}{@{}l@{}} $K$ for \\ $\topk{K}$  \\  \end{tabular} 
& \begin{tabular}{@{}l@{}} Sampled \\ from \end{tabular} 
& \begin{tabular}{@{}l@{}} Next \\ target \end{tabular} 
& \begin{tabular}{@{}c@{}} RMM \\ $\uparrow$ \end{tabular} 
& \begin{tabular}{@{}c@{}} Kinematic \\ metrics $\uparrow$ \end{tabular} 
& \begin{tabular}{@{}c@{}} Interactive \\ metrics $\uparrow$ \end{tabular} 
& \begin{tabular}{@{}c@{}} Map-based \\ metrics $\uparrow$ \end{tabular}
& \begin{tabular}{@{}c@{}} min \\ ADE $\downarrow$ \end{tabular}
\\ 
\cmidrule(lr){1-1}\cmidrule(lr){2-5}\cmidrule(lr){6-10}
BC pre-training &
- & - & - & GT
& $0.7581$ & $0.4512$ & $0.8076$ & $0.8697$ & $1.3152$ \\
\textbf{CAT-32 (submitted to leaderboard)} & prob & - & closest & GT 
& $\textbf{0.7616}$ & $\textbf{0.4583}$ & $\textbf{0.8105}$ & $0.8720$ & $\textbf{1.3105}$ \\
\cmidrule(lr){1-1}\cmidrule(lr){2-5}\cmidrule(lr){6-10}
\multirow{8}{*}{\begin{tabular}{@{}l@{}} Trajeglish's noisy \\ tokenization \end{tabular} }
& neg. dist. & 5 & neg. dist. & GT
& $0.7562$ & $0.4469$ & $0.8074$ & $0.8673$ & $1.3459$ \\
& neg. dist. & 5  & uniform & GT
& $0.7554$ & $0.4467$ & $0.8069$ & $0.8655$ & $1.3404$ \\
& neg. dist. & 16 & neg. dist. & GT
& $0.7486$ & $0.4336$ & $0.8031$ & $0.8585$ & $1.4811$ \\
& neg. dist. & 16  & uniform & GT
& $0.7481$ & $0.4315$ & $0.8033$ & $0.8581$ & $1.5012$ \\
& neg. dist. & 32 & neg. dist. &  GT
& $0.7401$ & $0.4174$ & $0.7985$ & $0.8493$ & $1.6669$ \\
& neg. dist. & 32 & uniform &  GT
& $0.7412$ & $0.4177$ & $0.7987$ & $0.8521$ & $1.6715$ \\
& neg. dist. & 64 & neg. dist. & GT
& $0.7303$ & $0.4005$ & $0.7906$ & $0.8413$ & $1.9083$ \\
& neg. dist. & 64  & uniform & GT
& $0.7295$ & $0.3994$ & $0.7890$ & $0.8416$ & $1.9307$ \\
\cmidrule(lr){1-1}\cmidrule(lr){2-5}\cmidrule(lr){6-10}
\multirow{8}{*}{\begin{tabular}{@{}l@{}} SMART's trajectory\\ perturbation \end{tabular} }
& neg. dist. & 5 & neg. dist. &  RO
& $0.7560$ & $0.4469$ & $0.8069$ & $0.8673$ & $1.3514$ \\
& neg. dist. & 5 & uniform & RO
& $0.7553$ & $0.4468$ & $0.8074$ & $0.8647$ & $1.3566$ \\
& neg. dist. & 16 & neg. dist. &  RO
& $0.7495$ & $0.4329$ & $0.8035$ & $0.8609$ & $1.4958$ \\
& neg. dist. & 16 & uniform & RO
& $0.7478$ & $0.4317$ & $0.8029$ & $0.8576$ & $1.4890$ \\
& neg. dist. & 32 & neg. dist. & RO
& $0.7407$ & $0.4190$ & $0.7985$ & $0.8503$ & $1.6472$ \\
& neg. dist. & 32 & uniform & RO
& $0.7403$ & $0.4179$ & $0.7986$ & $0.8497$ & $1.6568$ \\
& neg. dist. & 64 & neg. dist. &  RO
& $0.7309$ & $0.4012$ & $0.7917$ & $0.8411$ & $1.8701$ \\
& neg. dist. & 64 & uniform & RO
& $0.7284$ & $0.3962$ & $0.7879$ & $0.8417$ & $1.9574$ \\
\cmidrule(lr){1-10}
Top-16 & prob & 16 & prob & GT 
& $0.6439$ & $0.3309$ & $0.6912$ & $0.7619$ & $1.8744$ \\
Top-16 + distance filter & prob & 16 & prob & GT 
& $0.6904$ & $0.3375$ & $0.7489$ & $0.8169$ & $1.7991$ \\
Top-16 + distance based sampling & prob & 16 & neg. dist. & GT 
& $0.7233$ & $0.3675$ & $0.7808$ & $0.8528$ & $1.4876$ \\
\cmidrule(lr){1-1}\cmidrule(lr){2-5}\cmidrule(lr){6-10}
Top-32 & prob & 32 & prob & GT 
& $0.6395$ & $0.3324$ & $0.6882$ & $0.7522$ & $1.8961$ \\
Top-32 + distance filter & prob & 32 & prob & GT 
& $0.6950$ & $0.3400$ & $0.7560$ & $0.8193$ & $1.8194$ \\
Top-32 + distance based sampling & prob & 32 & neg. dist. & GT 
& $0.7229$ & $0.3663$ & $0.7843$ & $0.8477$ & $1.6470$ \\
\cmidrule(lr){1-1}\cmidrule(lr){2-5}\cmidrule(lr){6-10}
Top-64 & prob & 64 & prob & GT 
& $0.6381$ & $0.3318$ & $0.6846$ & $0.7535$ & $1.9117$ \\
Top-64 + distance filter & prob & 64 & prob & GT 
& $0.6979$ & $0.3407$ & $0.7590$ & $0.8234$ & $1.8172$ \\
Top-64 + distance based sampling & prob & 64 & neg. dist. & GT 
& $0.7208$ & $0.3660$ & $0.7823$ & $0.8446$ & $1.7260$ \\
\cmidrule(lr){1-1}\cmidrule(lr){2-5}\cmidrule(lr){6-10}
Trajeglish top-5, sampled w/ policy prob.
& neg. dist. & 5 & prob & GT
& $0.7596$ & $0.4513$ & $0.8089$ & $\textbf{0.8723}$ & $1.3116$ \\
Trajeglish top-32, sampled w/ policy prob.
& neg. dist. & 32 & prob &  GT
& $0.7526$ & $0.4320$ & $0.8069$ & $0.8659$ & $1.3569$ \\
SMART top-5, sampled w/ policy prob.
& neg. dist. & 5 & prob & RO
& $0.7589$ & $0.4510$ & $0.8085$ & $0.8709$ & $1.3135$ \\
SMART top-32, sampled w/ policy prob.
& neg. dist. & 32 & prob & RO
& $0.7580$ & $0.4533$ & $0.8093$ & $0.8661$ & $1.3325$ \\
\bottomrule
\end{tabular}
}
\caption{
\textbf{Ablation study on WOSAC 2\% validation split.}
We compare different ways to fine-tune the same base mode (BC pre-training). 
"Sampled from" indicates how the action is sampled during fine-tuning, either based on the distance to the GT (``neg. dist", ``uniform", ``closest") or based on the model outputs (``prob", ``max-prob").
Here dist. is the abbreviation of distance.
RO stands for rollout, i.e., the next target action is computed based on the rollout, not the GT state.
RMM stands for the realism meta metric of WOSAC.
}
\label{table:appendix_ablation}
\end{table*}
\subsection{Ablation}

In \cref{table:appendix_ablation}, we provide additional ablation studies we conducted.
Our method, CAT-K fine-tuning with $K=32$, achieves the overall best performance.
Only on the map-based metrics we are slightly outperformed by ``Trajeglish top-5, sampled w/ policy prob.'', but the difference is insignificant.
The ``sampled w/ policy prob.'' version of Trajeglish's noisy tokenization and SMART's trajectory perturbation is an on-policy variation of the original data augmentation, where the K closest-to-GT tokens are sampled using the probability predicted by the policy rather than using the negative distance.
These on-policy versions perform better than the off-policy data augmentation, but their performance is still worse than our CAT-K fine-tuning.
For top-K sampling, adding distance based filtering or distance based sampling improves the performance, but they still cannot match the performance of our method.
For the original versions of Trajeglish's and SMART's data augmentation, a thorough search of the hyperparameters confirms the conclusion drawn in the Trajeglish and SMART papers: Off-policy data augmentation does not significantly improve the performance on the WOSAC leaderboard.

Our CAT-K rollout can be seen as a special case of top-K with distance based sampling, where a very low temperature is used in the distance-based sampling, ensuring that the closest-to-GT token is selected deterministically.
For example, ``Top-32 + distance based sampling'' with a sampling temperature $\tau \to 0$ is equivalent to CAT-32 rollout.

\subsection{GMM-based ego policy}

In \cref{table:appendix_ego_policy} we present additional ablation studies for training and fine-tuning the GMM ego policy.
Inspired by the training strategy used in multimodal motion prediction, we experimented with applying hard-assignment to train the BC policy, aiming to mitigate the mode-averaging problem in the GMM.
Specifically, at each time step and for each agent, we train only the Gaussian mixture component that is closest to the GT, leaving the other components untrained.
However, this approach did not work, and the training diverged.
We then investigate the impact of Trajeglish's and SMART's data augmentation on fine-tuning the ego policy.
The results indicate that the effectiveness of these off-policy data augmentation methods is marginal: the RMM shows slight improvement, while the collision and off-road rates are marginally worse.
Next, we explore the use of top-K sampling for fine-tuning the BC policy.
As expected, top-K sampling alone does not work.
However, when combined with distance-based filtering or sampling, top-K sampling can significantly enhance the BC policy's performance, achieving results comparable to those of our CAT-K fine-tuning approach.
This justifies the effectiveness of this approach when the expert demonstrations are generally well-behaved and less diverse, which is consistent with prior work that applies this sampling strategy for fine-tuning to only vehicles~\cite{lu2023imitation}, often within the context of highway scenarios~\cite{zhang2023learning}.
Compared to top-K sampling with distance-based filtering or sampling, our method significantly outperforms in off-road rate and minADE, while other metrics remain on par.
Overall, fine-tuning with CAT-K rollout achieves the best performance, with peak performance at $K=2$ or $K=3$, which aligns with the fact that the ego vehicle's behavior is less multimodal.
For traffic simulations where demonstrations are highly multimodal and involve various traffic participants (vehicles, pedestrians, and cyclists) whose behaviors do not necessarily obey traffic rules, the advantage of our CAT-K rollout becomes more significant.

\begin{table*}
\setlength{\tabcolsep}{10pt}
\centering
\scalebox{\tablescaler}{
\begin{tabular}{lccccc} 
\toprule
\begin{tabular}{@{}l@{}} Method (\emph{Local val. split}) \end{tabular} 
& \begin{tabular}{@{}c@{}} Collision rate $\downarrow$ \end{tabular} 
& \begin{tabular}{@{}c@{}} Off-road rate $\downarrow$ \end{tabular} 
& \begin{tabular}{@{}c@{}} RMM $\uparrow$ \end{tabular} 
& \begin{tabular}{@{}c@{}} ADE $\downarrow$ \end{tabular} 
& \begin{tabular}{@{}c@{}} minADE$^{32}$ $\downarrow$ \end{tabular} 
\\
\cmidrule(lr){1-1}\cmidrule(lr){2-5}\cmidrule(lr){6-6}
BC pre-training &
$0.0568$ & $0.0053$ & $0.8108$ & $1.3623$  & $1.3537$ \\
\cmidrule(lr){1-1}\cmidrule(lr){2-5}\cmidrule(lr){6-6}
BC fine-tuning w/ hard-assignment (training diverged)  &
$0.1574$ & $0.0637$ & $0.7409$ & $5.3507$  & $5.3447$ \\
\cmidrule(lr){1-1}\cmidrule(lr){2-5}\cmidrule(lr){6-6}
Trajeglish noisy tokenization ($K=3$, neg. dist., GT) &
$0.0611$ & $0.0057$ & $0.8117$ & $1.3563$  & $1.3575$ \\
SMART trajectory perturbation ($K=3$, uniform, RO) &
$0.0590$ & $0.0057$ & $0.8118$ & $1.3713$  & $1.3771$ \\
\cmidrule(lr){1-1}\cmidrule(lr){2-5}\cmidrule(lr){6-6}
Top-3 &
$0.0415$ & $0.0140$ & $0.8072$ & $1.2004$  & $0.8249$ \\
Top-3 + distance filter &
$\mathbf{0.0409}$ & $0.0076$ & $0.8128$ & $1.1639$  & $0.8577$ \\
Top-3 + distance based sampling &
$0.0410$ & $0.0070$ & $0.8163$ & $1.3245$  & $0.7610$ \\
\cmidrule(lr){1-1}\cmidrule(lr){2-5}\cmidrule(lr){6-6}
CAT-1 (Deterministic rollout) &
$0.0433$ & $0.0138$ & $0.8081$ & $\mathbf{1.1799}$ & $0.7962$ \\
CAT-2 &
$0.0437$ & $0.0038$ & $0.8147$ & $1.5117$ & $\mathbf{0.6323}$ \\
CAT-3 &
$0.0422$ & $\mathbf{0.0035}$ & $\mathbf{0.8169}$ & $1.3096$ & $0.6912$ \\
CAT-4 &
$0.0500$ & $0.0035$ & $0.8137$ & $1.5699$ & $1.4840$ \\
CAT-8 &
$0.0771$ & $0.0045$ & $0.8050$ & $1.6775$ & $1.6704$\\
\bottomrule
\end{tabular}
}
\caption{\textbf{Performance of ego policies on WOSAC with local evaluation on 2\% validation split.}
All models are fine-tuned for 5 epochs based on the BC pre-training model, which is trained for 32 epochs.
We use deterministic rollout during inference and compute all metrics, except for the minADE$^{32}$.
For minADE$^{32}$, we generate 32 rollouts by using top-3 sampling with a temperature of $1.0$ to first sample the categorical distribution over the mixtures, then selecting the mean of the sampled Gaussian mixture.
RMM stands for the realism meta metric of WOSAC.
}
\label{table:appendix_ego_policy}
\end{table*}

\end{document}